%% file: acl_latex.tex
\pdfoutput=1

\documentclass[11pt]{article}

\usepackage[preprint]{acl}

\usepackage{latexsym}
\usepackage{wrapfig}
\input{math_commands.tex}

\usepackage{times}
\usepackage{color}
\usepackage{tabularx}
\usepackage{tabularray}

\usepackage[T1]{fontenc}

\usepackage[utf8]{inputenc}

\usepackage{microtype}

\usepackage{inconsolata}

\usepackage{graphicx}
\usepackage{microtype}

\usepackage{hyperref}       %
\usepackage{inconsolata}
\usepackage{algorithm}
\usepackage{wrapfig}
\usepackage{algorithmic}
\usepackage{colortbl}
\usepackage{wrapfig}
\usepackage{xcolor}
\usepackage{inconsolata}
\usepackage{multirow}  
\usepackage{amssymb}   
\usepackage{amsmath}
\usepackage{color}
\usepackage{fontawesome5}
\usepackage{subfigure} 
\usepackage{makecell} 
\usepackage{booktabs} 
\usepackage{enumitem} 
\usepackage{multirow} 

\usepackage{booktabs} 
\usepackage{stfloats}
\usepackage{listings}
\usepackage{cite}
\usepackage{pifont}
\usepackage{tcolorbox}
\usepackage{enumitem}
\tcbuselibrary{skins, breakable, theorems}
\usepackage{wrapfig}
\usepackage{lscape}
\usepackage{url}

\definecolor{mypink}{rgb}{.99,.91,.95}
\definecolor{myyellow}{rgb}{.99,.94,.82}

\input{config}

\title{GATEAU: 
Selecting Influential Samples for Long Context Alignment}
\author{
\textbf{Shuzheng Si\thanks{\ Equal Contribution.}$^{\spadesuit\diamondsuit}$, Haozhe Zhao\footnotemark[1]$^{\heartsuit}$, Gang Chen$^\diamondsuit$, Yunshui Li, Kangyang Luo$^{\spadesuit}$} \\
\textbf{{Chuancheng Lv$^\diamondsuit$}, Kaikai An$^{\heartsuit}$, Fanchao Qi\thanks{\ Corresponding Authors.}$^{\spadesuit\diamondsuit}$, Baobao Chang$^{\heartsuit}$,} and \textbf{Maosong Sun\footnotemark[2]$^{\spadesuit\bigstar}$} \\ 
$^{\spadesuit}$ Department of Computer Science and Technology, Tsinghua University \\ $^{\heartsuit}$ Peking University
\quad $^\diamondsuit$ DeepLang AI
\quad $^{\bigstar}$ Institute for AI, Tsinghua University
}

\begin{document}
\maketitle
\renewcommand{\thefootnote}{\fnsymbol{footnote}}
\renewcommand{\thefootnote}{\arabic{footnote}}

\input{Files/0_Abstract}
\input{Files/1_Introduction}

\input{Files/3_Method}

\input{Files/4_Experiment}

\input{Files/2_Related_work}

\input{Files/5_Conclusion}



\bibliography{custom}

\include{Files/6_Appendix}

\end{document}

%% file: math_commands.tex
\usepackage{amsmath,amsfonts,bm}

\def\eqref#1{equation~\ref{#1}}

\def\1{\bm{1}}

\DeclareMathAlphabet{\mathsfit}{\encodingdefault}{\sfdefault}{m}{sl}
\SetMathAlphabet{\mathsfit}{bold}{\encodingdefault}{\sfdefault}{bx}{n}



%% file: config.tex
\definecolor{Gray}{gray}{0.9}
\definecolor{mygreen}{rgb}{0.0, 0.5, 0.0}
\definecolor{myred}{rgb}{0.8, 0.25, 0.33}
\definecolor{myblue}{rgb}{0.19, 0.55, 0.91}
\definecolor{uclablue}{rgb}{0.15, 0.45, 0.68}
\definecolor{ucladblue}{rgb}{0.0, 0.33, 0.53}
\definecolor{ucladdblue}{rgb}{0.0, 0.23, 0.36}
\definecolor{uclagold}{rgb}{1.0, 0.82, 0.0}
\definecolor{ucladgold}{rgb}{1.0, 0.78, 0.17}
\definecolor{ucladdgold}{rgb}{1.0, 0.72, 0.11}
\definecolor{boxgreen}{rgb}{0.02, 0.66, 0.02}
\definecolor{boxred}{rgb}{0.66, 0.1, 0.1}
\definecolor{boxblue}{rgb}{0.01, 0.01, 0.73}

\usepackage{times}
\usepackage{latexsym}

\usepackage[utf8]{inputenc}
\usepackage[T1]{fontenc}

\usepackage{inconsolata}

\usepackage{algorithm}
\usepackage{algorithmic}

\usepackage{newfloat}
\usepackage{listings}
\floatstyle{ruled}
\newfloat{listing}{tb}{lst}{}
\floatname{listing}{Listing}

\usepackage{graphicx}
\usepackage{amsmath,amssymb,amsthm,mathabx,amsfonts}
\usepackage{bbm}
\usepackage{cleveref}
\usepackage{acronym}
\usepackage{enumitem}
\usepackage{balance}
\usepackage{xspace}
\usepackage{setspace}
\usepackage{url}
\usepackage{amsfonts}
\usepackage{multirow}
\usepackage{booktabs}
\usepackage{tablefootnote}
\usepackage[switch]{lineno}
\usepackage{multicol,lipsum}
\usepackage{tikz}
\usepackage{pgfplots}
\usepackage{pgfplotstable}
\usepackage{arydshln}

\usepackage{CJKutf8}

\pgfplotsset{compat=1.18}
\makeatletter
\DeclareRobustCommand\onedot{\futurelet\@let@token\@onedot}
\def\@onedot{\ifx\@let@token.\else.\null\fi\xspace}

\makeatother

\makeatletter
\newcommand{\thickhline}{%
    \noalign {\ifnum 0=`}\fi \hrule height 1pt
    \futurelet \reserved@a \@xhline
}

\crefname{algorithm}{Alg.}{Algs.}
\Crefname{algocf}{Algorithm}{Algorithms}
\crefname{section}{Sec.}{Secs.}
\Crefname{section}{Section}{Sections}
\crefname{table}{Tab.}{Tabs.}
\Crefname{table}{Table}{Tables}
\crefname{figure}{Fig.}{Fig.}
\Crefname{figure}{Figure}{Figure}
\crefname{appendix}{Appendix}{Appendices}

\acrodef{nlp}[NLP]{natural language processing}
\acrodef{plm}[PLM]{Pre-trained Language Model}
\acrodef{sota}[SOTA]{state-of-the-art}
\acrodef{icl}[ICL]{In-Context Learning}
\acrodef{bbl}[BBL]{BIG-bench Lite}

\usepackage{colortbl}
\definecolor{gblue}{HTML}{4285F4}
\definecolor{gred}{HTML}{DB4437}
\definecolor{ggreen}{HTML}{0F9D58}

\definecolor{mygray}{gray}{.92}
\definecolor{emphypurple}{rgb}{0.302, 0.055, 0.659}

\definecolor{highlightgreen}{HTML}{009901}
\definecolor{highlightred}{HTML}{FD6864}


%% file: Files/0_Abstract.tex
\begin{abstract}
Aligning large language models to handle instructions with extremely long contexts has yet to be fully investigated.
Previous studies have attempted to scale up the available data volume by synthesizing long instruction-following samples, as constructing such a dataset tends to be challenging for annotators.
However, a lack of a well-defined strategy for ensuring data quality may introduce low-quality samples and restrict the model’s performance.
Thus, we propose \textbf{GATEAU}, a novel framework to address the unique challenge of long context alignment by identifying the influential samples enriched with long-range dependency relations.
Specifically, GATEAU measures the long-range dependencies from two essential aspects: the difficulty of generating target responses due to the long-range dependencies, and the difficulty of understanding long inputs due to such dependencies.
Comprehensive experiments indicate that GATEAU effectively identifies influential samples and the model trained on these selected samples exhibits better instruction-following and long-context understanding capabilities.\footnote{~The data and code will be available at \url{https://github.com/S1s-Z/GATEAU}. Email: ssz24@mails.tsinghua.edu.cn.}

\end{abstract}

%% file: Files/1_Introduction.tex
\section{Introduction}
\label{section:introduction}
Large language models (LLMs) with large context
windows \citep{longchat2023, chen2024longlora} have shown impressive capabilities in real-world tasks that involve extremely long contexts \citep{bai2023longbench, si-etal-2022-mining, NEURIPS2023_7b16688a}.
Recent works to build long-context LLMs mainly focus on broadening context windows via position encoding extension and continual pre-training on a long corpus \citep{ peng2024yarn, xiong-etal-2024-effective}.

Despite these advancements, few studies consider the long context alignment of LLMs to leverage their capabilities in understanding lengthy inputs and following complex instructions.
A primary obstacle lies in the difficulty of constructing a high-quality, long instruction-following dataset for supervised fine-tuning (SFT). 
Annotating long instruction-following data tends to be much more challenging than short ones, as it is non-trivial for annotators to understand an excessively long context and provide high-quality responses, e.g., annotators might be tasked with writing a summary for a document containing 64k words.
Furthermore, modeling long-range dependencies is crucial for long-context tasks \citep{chen2024longcontextlongall, wu2024longcontextalignmentshort}, as such strong semantic dependencies benefit LLMs to understand lengthy inputs and generate high-quality responses.
Thus, recent works \citep{longchat2023,xiong-etal-2024-effective} attempt to construct the long instruction-following dataset by concatenating short instruction-following samples.
While these methods successfully increase sequence lengths, simply concatenating unrelated samples fails to effectively simulate the inherent long-range dependencies in authentic long samples.
To address this issue, \citet{yang2023longqlora,chen2024longlora, bai2024longalignrecipelongcontext} focus on synthesizing long instruction-following data.
For instance, \citet{bai2024longalignrecipelongcontext} synthesizes 10k samples by employing Claude 2.1 \citep{claude-2-1-2023}, which supports a context window of 200k tokens, to get responses for the collected long documents.

However, when training on such synthetic samples with sufficiently lengthy contexts, LLMs still struggle to learn and model the long-range dependencies \citep{chen2024longcontextlongall}.
This is because indiscriminately increasing the quantity of data without a well-defined strategy for ensuring data quality can introduce low-quality samples lacking long-range dependency relations.
Such samples may rely only on a few tokens before the instruction or may not require long inputs to get the target response.
Thus, a critical question arises: 
\textit{
\textbf{How can we effectively select influential samples from a vast amount of synthetic long instruction-following data for long context alignment?}
}

Previous studies for selecting influential instruction data primarily focus on short samples \citep{li-etal-2024-quantity,xia2024less}.
Thus, these studies may not be effective for long context alignment as they ignore the unique challenge in long context alignment, i.e., how to select the samples enriched with meaningful long-range dependency relations.
To address this challenge, we measure long-range dependencies from two essential aspects: the difficulty of generating target responses due to long-range dependencies, and the difficulty of understanding long inputs due to such dependencies.
We introduce \textbf{GATEAU}, which consists of \textbf{Homologous Models’ \underline{G}uid\underline{A}nce (HMG)} and \textbf{Con\underline{TE}xtual \underline{A}wareness Meas\underline{U}rement (CAM)}, to identify the influential long samples enriched with long-range dependency relations to achieve better long context alignment.

\begin{figure*}[t]
    \centering
    \includegraphics[width=0.881\linewidth]{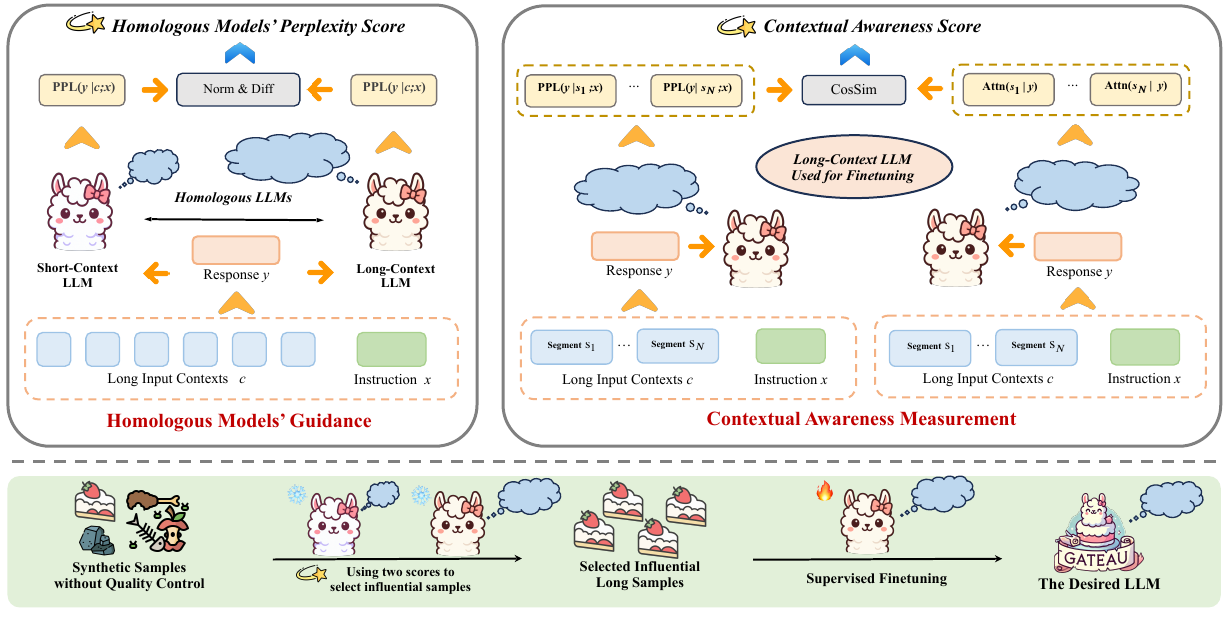}
    \caption{
    An overview of \textbf{GATEAU}. 
    GATEAU first selects samples enriched with long-range dependency relations by using two proposed methods.
    Then it uses selected influential samples for training long-context LLMs.
    }
    \label{figure:model}
\end{figure*}

Specifically, HMG measures the difficulty of generating target responses due to long-range dependencies,
by comparing perplexity scores of the given response between two homologous models \citep{yu2024language} with different context windows (e.g., the perplexity scores from LLaMA-3-base-8k \citep{llama3} and LLaMA-3-base-64k \citep{llama3-64k}).
The idea behind HMG is that the primary difference between homologous models with varying context windows lies in their different capabilities for modeling long-range dependencies.
Thus, the disparity in the perplexity scores can be interpreted as reflecting the difficulty of generating the response caused by long-range dependencies.
We further introduce CAM to measure the difficulty of understanding long input contexts due to long-range dependencies.
We first calculate the importance score of different input segments concerning the given response and subsequently measure whether LLMs can pay more attention to more important segments.
If LLM's attention focuses more on less important segments, it implies that it is challenging for the LLM to comprehend the long inputs correctly.
Ultimately, we take the weighted sum of both scores from the two methods as the final criterion for ranking the data, selecting the most challenging samples as influential ones.
When trained on these selected samples with rich long-range dependency relations, LLMs could effectively model the long-range dependencies and achieve better instruction-following performance.

We conduct extensive experiments to evaluate GATEAU, including long-context understanding benchmark (LongBench \citep{bai2023longbench}) and instruction-following benchmarks (LongBench-Chat \citep{bai2024longalignrecipelongcontext}, MT-Bench \citep{zheng2023judgingllmasajudgemtbenchchatbot}). 
With GATEAU, significant performance boosts are observed, e.g., the model trained on just 10\% selected samples of the dataset achieves better performance than the vanilla fine-tuning method.

%% file: Files/3_Method.tex
\section{Methodology}
\label{section:method}

As shown in Figure \ref{figure:model}, we propose \textbf{GATEAU} to select influential samples from a vast ocean of synthetic data instead of indiscriminately increasing the quantity of synthetic long instruction-following data \citep{chen2024longlora, bai2024longalignrecipelongcontext}.
Different from previous studies that only consider the short context scenarios \citep{li-etal-2024-quantity,xia2024less}, we attempt to address the unique challenge in long context alignment, i.e., modeling long-range dependencies.
GATEAU consists of Homologous Models’ Guidance and Contextual Awareness Measurement, which separately measure the difficulty of generating corresponding responses and understanding long input contexts due to the long-range dependencies.
In this way, GATEAU can comprehensively and effectively measure the richness of long-range dependency relations in long samples.

\subsection{Homologous Models’ Guidance}
Modeling long-range dependencies is essential for long context alignment \citep{chen2024longcontextlongall}.
However, there is still no effective metric to directly quantify the richness of long-range dependency relations in data, which hinders the selection of influential data.
Therefore, in this section, we attempt to approximately assess the richness of long-range dependency relations by measuring the difficulty in generating corresponding responses due to the long-range dependencies.
If LLMs find it harder to generate target responses due to long-range dependencies, it means the sample has more complex and meaningful long-range dependency relations.
An intuitive approach is to use the perplexity score to measure the difficulty of generating corresponding responses \citep{ cao2024instructionmininginstructiondata,li-etal-2024-quantity}, as the score evaluates the extent to which the LLM’s output aligns with the corresponding correct answer.
For a given long instruction-following sample $(c,x;y)$, the perplexity score of the given response $y$ from LLMs $\theta$ is calculated as:
\begin{align}
\scalebox{0.85}{
${\rm PPL}_{\theta}(y|c,x) = {\rm Exp}(-\frac{1}{|y|} \sum^{|y|}_{i=1} {\rm log} P(y_i|c,x,y_{<i} ; \theta ) ),$}
\label{equal-ppl}
\end{align}
where $c$ means long input contexts and $x$ means the given instruction.
A higher ${\rm PPL}_{\theta}(y|c,x)$ indicates the harder the response of this long instruction-following data for LLM to generate.

However, we argue that a higher ${\rm PPL}_{\theta}(y|x)$ does not mean the increased difficulty in generating target responses is due to long-range dependencies.
A higher ${\rm PPL}_{\theta}(y|c,x)$ might be attributed to certain limited capabilities of LLMs, such as the limited instruction-following capability for the model without alignment, instead of handling the long-range dependency relations in this sample is more challenging for the LLM.
Therefore, to minimize the influence of other factors, we propose \textbf{Homologous Models’ Guidance (HMG)}.
Specifically, we compare the perplexity scores of the response between two homologous models with different context windows to 
measure the difficulty due to the long-range dependencies.
As homologous models \citep{yu2024language} share the same pre-training stage and model architecture (e.g., 
LLaMA-3-base-8k and LLaMA-3-base-64k), the only difference lies in their capabilities to model long-range dependency due to the context windows extending stage.
Based on this motivation, we introduce the homologous models’ perplexity score $ {\rm HMP}(c,x;y)$:
\begin{align}  
{\rm HMP}(c,x;y) = {\rm Norm}({\rm PPL}_{\theta_A}(y|c,x)) \notag \\
- {\rm Norm}({\rm PPL}_{\theta_B}(y|c,x)). \label{eq_norm}  
\end{align}  

Model $\theta_A$ employs short context windows and $\theta_B$ is the model with long ones, e.g., LLaMA-3-base-8k $\theta_A$ and LLaMA-3-base-64k $\theta_B$.
We compute the difference in normalized perplexity scores between two homologous models with different context windows as the metric.
We apply softmax normalization to each score to determine its respective ranking among the datasets, since perplexity scores of one sample from different models often can't be directly compared.
By introducing a model $\theta_{A}$ with weaker long-range dependencies modeling capability but other similar capabilities learned during the pre-training stage, we mitigate the influence brought by lacking other capabilities compared to simply using the perplexity score as Eq. (\ref{equal-ppl}).
Thus, the difference in perplexity scores is primarily attributed to the different abilities in modeling long-range dependencies between model $\theta_A$ and model $\theta_B$.
In other words, Eq. (\ref{eq_norm}) reflects the difficulty of generating the corresponding response caused by long-range dependencies.
We use the drop from $\rm PPL_{\theta_A}$ to $\rm PPL_{\theta_B}$ in Eq. (\ref{eq_norm}) as model $\theta_A$ tends to produce a high perplexity score due to its weak ability to model long-range dependencies.
Thus, a higher $ {\rm HMP}(c,x;y)$ indicates more difficulties for LLM in response generation due to the long-range dependencies, i.e., more long-range dependency relations in this sample.

\subsection{Contextual Awareness Measurement}
Another challenge in long context alignment lies in enabling LLMs to understand and utilize extremely long inputs.
Due to the long-range dependencies, it is hard for LLMs to utilize crucial information hidden in extremely long contexts, e.g., LLM’s attention may focus on irrelevant content.
Thus, we introduce \textbf{Contextual Awareness Measurement (CAM)} to evaluate whether LLMs’ attention is appropriately focused on important segments within the lengthy inputs.
We attempt to evaluate the importance score of each segment and calculate the LLM’s attention weights on each one, getting the \textbf{Contextual Awareness Score (CAS)} via computing their similarity.
For a given data $(c,x;y)$, we divide the input contexts $c$ into $N$ segments $[s_1,s_2,s_3,...,s_N]$ of equal length $L$.
For segment $s_i$, we first compute the designed
importance score ${\rm IS}_{\theta}(s_i)$ to measure the significance of the segment in the response generation for LLM $\theta$:
\begin{align}
\scalebox{0.85}{
${\rm IS}_{\theta}(s_i) = {\rm Norm}({\rm Exp}(-\frac{1}{|y|} \sum^{|y|}_{j=1} {\rm log} P(y_i|s_i,x,y_{<j} ; \theta ) )).$}
\label{eq:is}
\end{align}

We only keep the segment $s_i$ as the input to calculate the perplexity score of generating the response $y$, indicating the difficulty of generating response $y$ based on segment $s_i$.
We apply softmax normalization to each score 
to determine its respective ranking among the segments $\{s_i\}_{i=1}^{N}$ as shown in Eq. (\ref{eq:is}).
The higher ${\rm IS}_{\theta}(s_i)$ suggests a greater difficulty for LLM $\theta$ to generate the response based on segment $s_i$, implying that it is less important.

Once the importance scores of different segments are calculated, we then utilize the attention weights (i.e., the value of $ {\rm softmax}(\frac{QK^T}{\sqrt{d_k}})$ ) in the multi-head attention mechanism \citep{Vaswani2017AttentionIA} to measure how the LLM utilizes these segments.
We use the averaged attention weights of tokens $[t_1,...,t_{L}]$ in segments $s_i$ as the score $ {\rm Attn}_{\theta}(s_i)$, which takes the form:
\begin{align}
{\rm Attn}_{\theta}(s_i) = {\rm Norm}(\frac{1}{L} \sum^{L}_{j=1} {\rm Attn}_{\theta}(t_j|y;\theta)),
\end{align}
where $ {\rm Attn}_{\theta}(t_j|y;\theta)$ means the attention weights averaged across the tokens in targeted response $y$ to the token $t_j$ in segment $s_i$.
Meanwhile, we harness the attention weights averaged across different decoder layers and attention heads to thoroughly model how the LLM utilizes the long input contexts during the response generation \citep{hsieh2024middlecalibratingpositionalattention}.
We apply softmax normalization to each score $\frac{1}{L}\sum_{j=1}^{L} {\rm Attn}_{\theta}(t_j|y;\theta)$ to determine its respective ranking among the segments $\{s_i\}_{i=1}^{N}$ to yield
the score ${\rm Attn}_{\theta}(s_i)$.
In so doing, we can calculate the attention weights between the response and segments, indicating how segments are utilized during the response generation.

Finally, we can measure the difficulty of understanding the long input contexts due to the long-range dependencies.
For a given long instruction-following sample,
we compute the CAS by resorting to the cosine similarity between importance scores $[{\rm IS}_\theta(s_1),..., {\rm IS}_\theta(s_N)]$ and attention weights $[{\rm Attn}_\theta(s_1),..., {\rm Attn}_\theta(s_N)]$, as follows:
\begin{align}  
{\rm CAS}(c,x;y) = &{\rm CosSim}([{\rm IS}_\theta(s_1), \ldots, {\rm IS}_\theta(s_N)], \notag \\
&[{\rm Attn}_\theta(s_1), \ldots, {\rm Attn}_\theta(s_N)]). \label{cas_eq}  
\end{align}

By doing this, we can measure the difficulty of understanding the long input contexts by evaluating whether LLMs’ attention is focused on important segments.
The insight is that if the LLM’s attention focuses more on less important segments, it suggests that the LLM struggles to accurately comprehend the given long input contexts.
The higher ${\rm CAS}(c,x;y)$ indicates more difficulties in utilizing the long input contexts to generate corresponding responses due to the long-range dependencies.

\subsection{Selecting and Training}
We frame the final score by weighting two metrics of the sample $(c,x;y)$, then select the most challenging samples as the influential samples, i.e.,
\begin{align}  
{\rm Score}(c,x;y) = \alpha*{\rm Norm}({\rm HMP}(c,x;y)) \notag \\
+ (1 - \alpha)*{\rm Norm}({\rm CAS}(c,x;y)), 
\label{final_score}  
\end{align}
where $\alpha$ is a hyperparameter.
We tap softmax normalization to the ${\rm HMP}(c,x;y)$ and ${\rm CAS}(c,x;y)$ across the whole dataset.
Inspired by active learning \citep{li2024surveydeepactivelearning}, when trained on these challenging data with complex long-range dependency relations, LLMs could learn such dependencies and achieve better long context alignment.

Training LLMs with instruction-following data can teach LLMs to follow user instructions.
Thus, we apply SFT on the selected data (e.g., selecting 10\% samples of full datasets with top 10\% scores according to Eq. (\ref{final_score})).
Then, we train LLMs using the following objective function:
\begin{align}
\mathcal{L}_{\theta}(c,x;y) = - \sum^{|y|}_{i=1} {\rm log} P(y_i|c,x,y_{<i} ; \theta ).
\end{align}

%% file: Files/4_Experiment.tex
\section{Experiment}
\label{section:experiment}

\subsection{Experimental Setup}
\label{sec_setup}

\textbf{Training Datasets.}
We use LongAlign \citep{bai2024longalignrecipelongcontext} as the long instruction-following dataset, which contains 10,000 long samples.
LongAlign is developed by using collected long sequences from 9 sources and applying the Self-Instruct \citep{wang-etal-2023-self-instruct} approach with long-context LLM Claude 2.1 \citep{claude-2-1-2023}.
Though initially competitive, its dependence on Claude 2.1 synthesized data may lead to quality concerns.
Thus, our method to apply the selection of long instruction data is based on the LongAlign dataset.
We apply GATEAU to the LongAlign dataset.
Meanwhile, similar to \citet{bai2024longalignrecipelongcontext}, 
to maintain the model’s general capabilities and its proficiency in following short instructions, we utilize the ShareGPT dataset \citep{sharegpt} as the source of short instruction data in training data. 
To study the impact of mixing long and short instruction samples, we evaluate GATEAU in both \textbf{Real-world Settings} and \textbf{Limited Short Instruction Data Settings}.
Real-world Settings indicate real-world users prioritize short instruction-following interactions \citep{sharegpt}.
Thus, we use the full ShareGPT dataset as short instruction-following data.
We also explore scenarios where short instruction data is limited, utilizing only the first 10\% of ShareGPT, namely Limited Short Instruction Data Settings.
More details are shown in the Appendix \ref{appendix:training}.

\noindent
\textbf{Training Details.}
For the fair comparisons with \citet{bai2024longalignrecipelongcontext}, we use LLaMA-2-7B-base-4k \citep{touvron2023llama2openfoundation} and LLaMA-2-7B-base-64k \citep{bai2024longalignrecipelongcontext} as homologous models to apply HMG.
For CAM, we use LLaMA-2-7B-base-64k to calculate the CAS.
We train the LLaMA-2-7B-base-64k based on selected samples as our final model GATEAU-LLaMA.
We also find GATEAU can fit in other LLMs in the Appendix \ref{appendix:dis_other}, including ChatGLM-3 \citep{zeng2023glmb, bai2024longalignrecipelongcontext} and LLaMA-3 series \citep{llama3,llama3-64k}.
More details are shown in the Appendix \ref{appendix:training}.

\noindent
\textbf{Baselines.}
We compare our method with multiple SFT data selection baselines.
\textbf{Cherry Selection} \citep{li-etal-2024-quantity} and \textbf{CaR} \citep{ge2024clustering} are state-of-the-art methods to select the influential short instruction-following data.
We also use the perplexity score from long-context LLM as guidance to select long instruction-following samples according to Eq. (\ref{equal-ppl}), namely \textbf{Perplexity Guidance}.
More details can be found in the Appendix \ref{appendix:baselines}.

\begin{table}
\scriptsize	
\centering
\resizebox{0.8\linewidth}{!}{
\begin{tabular}{lcc}
\\
\toprule
\textbf{Model} & \textbf{Real-world} & \textbf{Limited}\\
\midrule
\multicolumn{3}{c}{\cellcolor{myyellow} \textbf{LongBench-Chat}}\\
w/o SFT & 10.4 & 10.4 \\
w/o Long SFT& 37.4 & 36.2 \\
Full - 100\% & 48.8 & 50.8 \\
\hline
Perplexity Guidance - 10\%	& 52.2 & 49.0
\\
CaR -  10\%	& 50.8 & 49.0 \\
Cherry Selection - 10\%	& 53.2 & 50.8 \\
\rowcolor{blue!5} \textbf{GATEAU-LLaMA - 10\%}& \textbf{55.4}	& \textbf{58.0} \\
\hline
Perplexity Guidance - 30\%	& 50.6
 & 51.8 \\
CaR -  30\%	& 48.6 & 51.4 \\
Cherry Selection - 30\%	& 50.4 & 52.4 \\
\rowcolor{blue!5} \textbf{GATEAU-LLaMA - 30\%}& \textbf{57.8}	& \textbf{55.2} \\
\hline
Perplexity Guidance - 50\%	& 49.8
 & 51.0 \\
CaR -  50\%	& 49.6 & 51.6 \\
Cherry Selection - 50\%	& 50.6 & 53.2 \\
\rowcolor{blue!5} \textbf{GATEAU-LLaMA - 50\%}& \textbf{56.8}	& \textbf{59.0}
\\
\midrule
\multicolumn{3}{c}{\cellcolor{myyellow} \textbf{MT-Bench}}\\
w/o SFT & 34.6
 &  34.6\\
w/o Long SFT & 53.7 & 50.5 \\
Full - 100\% &  54.3
 & 47.7 \\
\hline
Perplexity Guidance - 10\%	& 56.1
 & 50.9
\\
CaR -  10\%	& 54.9
 & 49.9
 \\
Cherry Selection - 10\%	&  56.8
&  47.6
\\
\rowcolor{blue!5} \textbf{GATEAU-LLaMA - 10\%} & \textbf{58.6}
	& \textbf{53.4} \\
\hline
Perplexity Guidance - 30\%	& 55.0
 &  50.2 \\
CaR -  30\%	& 54.3
 & 48.6 \\
Cherry Selection - 30\%	& 54.3
 & 45.8 \\
\rowcolor{blue!5} \textbf{GATEAU-LLaMA - 30\%}& 	\textbf{58.8}
& \textbf{52.9} \\
\hline
Perplexity Guidance - 50\%	& 55.9 & 49.2
\\
CaR -  50\%	&  54.7
&  51.2 \\
Cherry Selection - 50\%	& 56.3
 &  49.6 \\
\rowcolor{blue!5} \textbf{GATEAU-LLaMA - 50\%}& \textbf{57.3}
	& \textbf{54.2} \\
\bottomrule
\end{tabular}}
\caption{Results (\%) on LongBench-Chat and MT-Bench in two different settings.}
\label{tb:longchat_mt} 
\end{table}

\noindent
\textbf{Evaluation.}
To gauge the effectiveness of our method, we conduct extensive evaluations on different benchmarks.
We use \textbf{LongBench-Chat} \citep{bai2024longalignrecipelongcontext} to evaluate the models’ ability to follow long instructions, which comprises open-ended questions of 10k-100k in length.
We also employ a bilingual and multi-task benchmark \textbf{LongBench} \citep{bai2023longbench} to evaluate the model’s long-context understanding abilities.
We conduct evaluations on three tasks following \citet{bai2024longalignrecipelongcontext}, including Single-Doc QA, Multi-Doc QA, and Summarization.
Meanwhile, as aligned models generally produce longer responses, rather than relying solely on the automatic metrics (e.g., ROUGE) to evaluate the results, we follow \citet{bai2024longalignrecipelongcontext} to employ GPT-4 to evaluate the model outputs based on their alignment with the ground-truth answers on LongBench. 
We use \textbf{MT-Bench} \citep{zheng2023judgingllmasajudgemtbenchchatbot} to measure the models’ ability to follow short instructions via GPT-4 rating.
To ensure the most stable evaluation results, we use GPT-4 to score twice and average these scores to obtain the final results.
More details about evaluation (e.g., the rating prompts) can be found in the Appendix \ref{appendix:evaluations}.

\begin{table*}[t]
\scriptsize
\centering  
\resizebox{\textwidth}{!}{
\begin{tabular}{lccccccccccccccc}
\toprule
\multirow{2}{*}{\textbf{Model}} & \multicolumn{5}{c}{\textbf{Single-Doc QA}} & \multicolumn{5}{c}{\textbf{Multi-Doc QA}} & \multicolumn{5}{c}{\textbf{Summarization}} \\
\cmidrule(lr){2-6} \cmidrule(lr){7-11} \cmidrule(lr){12-16} 
& \textbf{1-1} & \textbf{1-2} & \textbf{1-3} & \textbf{1-4} & \textbf{\textcolor[rgb]{0.8,0,0}{Avg}} & \textbf{2-1} & \textbf{2-2} & \textbf{2-3} & \textbf{2-4} & \textbf{\textcolor[rgb]{0.8,0,0}{Avg}} & \textbf{3-1} & \textbf{3-2} & \textbf{3-3} & \textbf{3-4} & \textbf{\textcolor[rgb]{0.8,0,0}{Avg}} \\
\midrule
\rowcolor{mygray} \multicolumn{16}{c}{\cellcolor{mypink} \textbf{Real-world Settings}} \\
w/o SFT	&33.8 	&38.0 &	41.1 &	34.8 &	36.9 	&41.3 &	37.2 	&33.3& 	42.0 	&38.5 	&39.2 &	20.2 	&37.1 	&30.9 &	31.9 \\
w/o Long SFT	&58.7 &	66.7 &	83.1& 	79.2 	&71.9 &	70.2 &	53.4 &	48.7 	&61.3 &	58.4 	&57.3 &	36.2 	&55.2 &	38.4 &	46.8 \\
Full - 100\%	&62.8& 	69.0 	&83.1 &	81.3 	&74.1 	&71.5 	&54.8 	&51.3 &	66.2 	&61.0 	&58.7 	&39.8 	&57.6 &	41.2 	&49.3  \\
\midrule
Perplexity Guidance - 10\%	&62.0 &	68.8 &	86.4 &	85.6 &	75.7 &	73.5 	&59.7 &	52.1 	&68.2 &	63.4 &	67.6  &	41.3 	&\textbf{67.0} &	44.9 	&55.2  \\
CaR -  10\%	& 60.3 &	69.0 	&86.0 	&84.8 &	75.0 &	69.1 	&58.3 	&52.3 &	68.5 	&62.1 	&64.1 &	41.4 &	60.3 &	42.1 	&52.0 \\
Cherry Selection - 10\%	&60.8 	&67.2 	&86.7 	&84.3 	&74.8 	&71.3 &	57.8 &	51.0 &	69.0 	&62.3 &	61.3 &	40.0 &	64.8 	&41.5 &	51.9 \\
 \rowcolor{blue!5} \textbf{GATEAU-LLaMA - 10\%} & \textbf{63.6} 	&\textbf{69.2} &	\textbf{86.9} &	\textbf{87.1} &	\textbf{76.7} &	\textbf{74.8}&	\textbf{60.8}	&\textbf{53.1}&	\textbf{69.5}&	\textbf{64.6} &	\textbf{67.6}&	\textbf{42.6}&	66.2&	\textbf{47.8}&	\textbf{56.1} \\
\rowcolor{blue!5}  $\Delta$ compared to Full - 100\%  &  \textcolor[rgb]{0.7,0,0}{+0.8}  &	\textcolor[rgb]{0.7,0,0}{+0.2}  &	\textcolor[rgb]{0.7,0,0}{+3.8} 	 &\textcolor[rgb]{0.7,0,0}{+5.8}  &	\textcolor[rgb]{0.7,0,0}{+2.7}  &	\textcolor[rgb]{0.7,0,0}{+3.3} 	 &\textcolor[rgb]{0.7,0,0}{+6.0}  &	\textcolor[rgb]{0.7,0,0}{+1.8}  &	\textcolor[rgb]{0.7,0,0}{+3.3}  &	\textcolor[rgb]{0.7,0,0}{+3.6} 	 &\textcolor[rgb]{0.7,0,0}{+8.9}  &	\textcolor[rgb]{0.7,0,0}{+2.8}  &	\textcolor[rgb]{0.7,0,0}{+8.6} 	 &\textcolor[rgb]{0.7,0,0}{+6.6}  &	\textcolor[rgb]{0.7,0,0}{+6.7} \\
\midrule
Perplexity Guidance -  30\%	&62.8 &	67.3 &	86.2 &	82.6 &	74.7 	&72.3 &	59.3 &	50.8 	&67.8 &	62.6 &	62.3 &	41.7 	&64.8 &	42.7 &	52.9 \\
CaR -  30\%	&61.3 	&67.3 &	86.4 	&85.3 &	75.1 	&68.3 &	58.3 	&53.2 	&66.8 	&61.7 &	64.6 &	39.7 	&60.7 &	41.2 &	51.6 \\
Cherry Selection - 30\%	&62.0 	&66.8 &	87.1 &	84.3 &	75.1 &	74.3 &	59.3 	&52.7 	&68.7 	&63.8 &	62.3 &	40.5 &	64.6 	&44.4 &	53.0 \\
\rowcolor{blue!5} \textbf{GATEAU-LLaMA - 30\%}	& \textbf{63.0} &	\textbf{70.8} 	& \textbf{87.6} &	\textbf{85.8} &	\textbf{76.8} &	\textbf{75.7} 	& \textbf{61.0} &	\textbf{55.7} 	& \textbf{69.5} &	\textbf{65.5} &	\textbf{67.5} 	& \textbf{44.7} 	& \textbf{65.9} &	\textbf{47.4} 	& \textbf{56.4} \\
\rowcolor{blue!5}  $\Delta$ compared to Full - 100\%  &  \textcolor[rgb]{0.7,0,0}{+0.2} &	\textcolor[rgb]{0.7,0,0}{+1.8} 	&\textcolor[rgb]{0.7,0,0}{+4.5} 	&\textcolor[rgb]{0.7,0,0}{+4.5} 	&\textcolor[rgb]{0.7,0,0}{+2.8} 	&\textcolor[rgb]{0.7,0,0}{+4.2} &	\textcolor[rgb]{0.7,0,0}{+6.2} &	\textcolor[rgb]{0.7,0,0}{+4.4} &	\textcolor[rgb]{0.7,0,0}{+3.3} &	\textcolor[rgb]{0.7,0,0}{+4.5} &	\textcolor[rgb]{0.7,0,0}{+8.8} &	\textcolor[rgb]{0.7,0,0}{+4.9}& 	\textcolor[rgb]{0.7,0,0}{+8.3} &	\textcolor[rgb]{0.7,0,0}{+6.2} 	&\textcolor[rgb]{0.7,0,0}{+7.1}  \\
\midrule
Perplexity Guidance - 50\%	&63.1 &	68.1 	&87.8 &	82.1 &	75.3 	&74.2 &	59.2 &	52.5 	&69.2 &	63.8 &	64.7 &	41.1 &	65.7 &	42.1 &	53.4 \\
CaR -  50\% &	60.0 	&66.3 &	85.6 &	84.2 	&74.0 &	70.7 	&55.8& 	54.3 &	68.2 &	62.3 	&64.4 &	41.1 & 60.8 &	40.3 &	51.7 \\
Cherry Selection - 50\% &	62.8 	&65.5 &	86.2 &	82.8 	&74.3& 	72.2 &	56.8 &	52.7 	&67.8& 	62.4 &	64.6 &	39.4 	&64.1 &	42.1 &	52.6 \\
\rowcolor{blue!5} \textbf{GATEAU-LLaMA - 50\%}& \textbf{63.5} &	\textbf{70.3} &	\textbf{89.7} &	\textbf{86.5} &	\textbf{77.5} 	& \textbf{75.3}	& \textbf{60.8}	& \textbf{53.5} &	\textbf{68.5}	& \textbf{64.5} &	\textbf{65.1}	& \textbf{41.6}	& \textbf{65.9}	& \textbf{46.1}	& \textbf{54.7}  \\
\rowcolor{blue!5}  $\Delta$ compared to Full - 100\%  & \textcolor[rgb]{0.7,0,0}{+0.7} 	 &\textcolor[rgb]{0.7,0,0}{+1.3} 	 &\textcolor[rgb]{0.7,0,0}{+6.6}  &	\textcolor[rgb]{0.7,0,0}{+5.2}  &	\textcolor[rgb]{0.7,0,0}{+3.5} 	 &\textcolor[rgb]{0.7,0,0}{+3.8}  &	\textcolor[rgb]{0.7,0,0}{+6.0}  &	\textcolor[rgb]{0.7,0,0}{+2.2}  &	\textcolor[rgb]{0.7,0,0}{+2.3} 	 &  \textcolor[rgb]{0.7,0,0}{+3.6}  &	\textcolor[rgb]{0.7,0,0}{+6.4}  &	\textcolor[rgb]{0.7,0,0}{+1.8}  &	\textcolor[rgb]{0.7,0,0}{+8.3} 	 & \textcolor[rgb]{0.7,0,0}{+4.9} 	 & \textcolor[rgb]{0.7,0,0}{+5.4}  \\
\midrule
\rowcolor{mygray} \multicolumn{16}{c}{\cellcolor{mypink} \textbf{
Limited Short Instruction Data Settings}} \\
w/o SFT	&33.8 	&38.0 &	41.1 &	34.8 &	36.9 	&41.3 &	37.2 	&33.3& 	42.0 	&38.5 	&39.2 &	20.2 	&37.1 	&30.9 &	31.9 \\
w/o Long SFT &  62.3 &	70.8 &	88.5 	&82.7&	76.1 &	72.8	&60.6&	51.8	&67.3	&63.1 & 64.7 	&41.1& 	61.4 &	41.6 	&52.2 \\
Full - 100\%	&58.7 &	69.7& 	85.8 &	83.0 &	74.3 &	70.5 &	58.7 &	50.8 &	67.8 &	62.0 &	59.6 	&38.4 	&59.6 &	43.3 &	50.2  \\
\midrule
Perplexity Guidance - 10\%	&62.8 &	69.2 &	89.3 	&85.7 	&76.8 &	73.8 &	59.1 	&54.1 &	\textbf{71.1} &	64.5 &	69.8 	&45.8 &	65.7 	&50.1 &	57.9  \\
CaR -  10\%	& 62.8& 	68.3 &	88.0 	&82.7 &	75.5 &	71.8 &	58.0 &	52.7 &	68.8 &	62.8 &	65.5 &	42.0 &	61.8 	&43.1 &	53.1 \\
Cherry Selection - 10\%	&62.8 &	69.8 &	86.7 &	85.7 &	76.3 	&72.0 &	58.7 	&52.5 &	69.3 	&63.1 &	63.2 &	43.3 &	60.1 	&46.4 &	53.3  \\
\rowcolor{blue!5} \textbf{GATEAU-LLaMA - 10\%}	& \textbf{64.8} 	&\textbf{74.7} &	\textbf{89.8} 	&\textbf{86.5} &	\textbf{79.0} &	\textbf{75.2} &	\textbf{61.2} &	\textbf{54.6} &	70.0 &	\textbf{65.3} &	\textbf{71.1} &	\textbf{47.3} &	\textbf{67.0} &	\textbf{54.2} 	& \textbf{59.9}  \\
\rowcolor{blue!5}  $\Delta$ compared to Full - 100\%  &  \textcolor[rgb]{0.7,0,0}{+6.1} 	&  \textcolor[rgb]{0.7,0,0}{+5.0} 	&  \textcolor[rgb]{0.7,0,0}{+4.0} 	&  \textcolor[rgb]{0.7,0,0}{+3.5} 	&  \textcolor[rgb]{0.7,0,0}{+4.7} 	&  \textcolor[rgb]{0.7,0,0}{+4.7} 	&  \textcolor[rgb]{0.7,0,0}{+2.5} 	&  \textcolor[rgb]{0.7,0,0}{+3.8} 	&  \textcolor[rgb]{0.7,0,0}{+2.2} 	&  \textcolor[rgb]{0.7,0,0}{+3.3} 	&  \textcolor[rgb]{0.7,0,0}{+11.5} 	&  \textcolor[rgb]{0.7,0,0}{+8.9} 	&  \textcolor[rgb]{0.7,0,0}{+7.4} 	&  \textcolor[rgb]{0.7,0,0}{+10.9} 	&  \textcolor[rgb]{0.7,0,0}{+9.7} \\

\midrule
Perplexity Guidance -  30\%	&62.5 &	71.8 &	88.2 &	83.8& 	76.6 &	74.6 	&58.5 &	53.5 &	69.3 &	64.0& 	67.5 &	44.0 &	64.7 &	50.4 &	56.7 \\
CaR -  30\%	&60.8 	 &70.7 & 	88.4  &	81.8  &	75.4  &	73.0  &	59.0 	 &53.5  &	68.5 	 &63.5  &	64.1  &	40.9  &	62.3 	 &45.8  &	53.3  \\
Cherry Selection - 30\%	&62.8 &	71.7 &	88.9 &	\textbf{87.5} &	77.7 	&70.3 	&58.7 &	50.3& 	68.2 &	61.9 	&62.9 	&43.5 &	65.2 &	44.6 &	54.1 \\
\rowcolor{blue!5} \textbf{GATEAU-LLaMA - 30\%}	& \textbf{64.8} 	& \textbf{73.0} 	& \textbf{89.3} &	86.2 &	\textbf{78.3} &	\textbf{74.7} &	\textbf{61.0} 	& \textbf{54.2} &	\textbf{69.8} &	\textbf{64.9} &	\textbf{70.8} 	&\textbf{46.0} &	\textbf{66.4} 	& \textbf{51.4} 	& \textbf{58.7} \\
\rowcolor{blue!5}  $\Delta$ compared to Full - 100\%  &  \textcolor[rgb]{0.7,0,0}{+6.1} 	&  \textcolor[rgb]{0.7,0,0}{+3.3} 	&  \textcolor[rgb]{0.7,0,0}{+3.5} 	&  \textcolor[rgb]{0.7,0,0}{+3.2} 	&  \textcolor[rgb]{0.7,0,0}{+4.0} 	&  \textcolor[rgb]{0.7,0,0}{+4.2} 	&  \textcolor[rgb]{0.7,0,0}{+2.3} 	&  \textcolor[rgb]{0.7,0,0}{+3.4} 	&  \textcolor[rgb]{0.7,0,0}{+2.0} 	&  \textcolor[rgb]{0.7,0,0}{+3.0} 	&  \textcolor[rgb]{0.7,0,0}{+11.2} 	&  \textcolor[rgb]{0.7,0,0}{+7.6} 	&  \textcolor[rgb]{0.7,0,0}{+6.8} 	&  \textcolor[rgb]{0.7,0,0}{+8.1} 	&  \textcolor[rgb]{0.7,0,0}{+8.4} \\

\midrule
Perplexity Guidance - 50\%	&61.5 &	68.3 &	85.1 &	82.8 &	74.4 &	72.3 	&59.3 &	52.0 &	67.7 &	62.8 &	60.2 &	40.9 &	58.6 &	42.3 &	50.5 \\
CaR -  50\% &	62.3  &		68.1  &		86.9  &		80.1  &		74.4  &		71.0  &		58.7  &		52.8  &		68.0  &		62.6  &		64.4  &		41.2 	 &	61.1  &		45.6  &		53.1  \\
Cherry Selection - 50\% & 61.2 &	69.7 &	86.2 	& 83.7 	&75.2& 	69.7 &	56.8 &	49.5 	&66.2 &	60.6 &	64.1 &	41.8 &	60.5 	&43.7 	&52.5  \\
\rowcolor{blue!5} \textbf{GATEAU-LLaMA - 50\%}&	\textbf{63.7} 	& \textbf{71.8} &	\textbf{87.1} &	\textbf{84.7} &	\textbf{76.8} 	& \textbf{74.0} 	& \textbf{60.0} 	& \textbf{53.8} 	& \textbf{69.0} 	& \textbf{64.2} & 	\textbf{66.1} &	\textbf{43.9} &	\textbf{62.4} 	& \textbf{46.4} &	\textbf{54.7}  \\
\rowcolor{blue!5}  $\Delta$ compared to Full - 100\%  &  \textcolor[rgb]{0.7,0,0}{+5.0} &  \textcolor[rgb]{0.7,0,0}{+2.1} 	&  \textcolor[rgb]{0.7,0,0}{+1.3} 	&  \textcolor[rgb]{0.7,0,0}{+1.7} 	&  \textcolor[rgb]{0.7,0,0}{+2.5} 	&  \textcolor[rgb]{0.7,0,0}{+3.5} 	&  \textcolor[rgb]{0.7,0,0}{+1.3} 	&  \textcolor[rgb]{0.7,0,0}{+3.0} 	&  \textcolor[rgb]{0.7,0,0}{+1.2} 	&  \textcolor[rgb]{0.7,0,0}{+2.3} 	&  \textcolor[rgb]{0.7,0,0}{+6.5} 	&  \textcolor[rgb]{0.7,0,0}{+5.5} 	&  \textcolor[rgb]{0.7,0,0}{+2.8} 	&  \textcolor[rgb]{0.7,0,0}{+3.1} 	&  \textcolor[rgb]{0.7,0,0}{+4.5} \\
\bottomrule
\end{tabular}
}
\caption{GPT-4 evaluation results (\%) on LongBench in Real-world Settings. 
We use the ID to represent the dataset in LongBench, e.g., 1-1 is the ID of the NarrativeQA dataset. 
More details can be found in the Appendix \ref{appendix_longbench}.
Automatic metrics evaluation results (\%) are shown in Table \ref{tb:longbench-auto}.
}
\label{tb:longbench-gpt4}
\end{table*}

\subsection{Main Results}

\noindent
\textbf{GATEAU Improves Instruction-Following Capabilities for Both Short and Long Inputs.}
The results are shown in 
Table \ref{tb:longchat_mt} for the LongBench-Chat and MT-Bench benchmarks in two settings.
It shows that GATEAU can improve LLMs’ capabilities in following both long and short instructions and generating high-quality responses.
Compared to indiscriminately using the whole dataset, using the selected subset of the long instruction-following dataset (\textit{GATEAU-LLaMA}) can significantly improve the instruction-following capabilities, e.g., increasing 9\% in LongBench-Chat and 6.5\% in MT-Bench. 
Meanwhile, the low performance of \textit{w/o Long SFT} in LongBench-Chat indicates that using long SFT data is important for the performance in handling the instructions with long input contexts.
The results also show that our method GATEAU achieves consistently better performance in varying ratios of used long instruction-following samples compared with other baselines, indicating the effectiveness of our method.
Compared with baselines focusing on short SFT samples (\textit{CaR} and \textit{Cherry Selection}), GATEAU can identify samples enriched with long-range dependency relations more effectively and help LLMs to achieve better long context alignment.
We also observe that the selection of long instruction-following samples aids in augmenting the instruction-following capabilities for short inputs. 
We conjecture that handling complex tasks (i.e., long input contexts) contributes to handling the easy ones (i.e., short input contexts).

\noindent
\textbf{GATEAU Enhances the Long-Context Understanding Capabilities.}
\
The results are shown in Table \ref{tb:longbench-gpt4} and Table \ref{tb:longbench-auto} (in the Appendix) for the LongBench benchmark.
Our methods achieve consistent and remarkable performance gains in different settings and evaluation methods.
We show the improved scores (\textit{$\Delta$ compared to Full-100\%}) compared to indiscriminately using the whole dataset (\textit{Full-100\%}), indicating that GATEAU
helps LLM to better understand the long input contexts.
We also find that the baselines focusing on the selection of short instruction-following data (\textit{CaR} and \textit{Cherry Selection}) hold inferior results, sometimes even worse than using the whole dataset (\textit{Full-100\%}).
This can be attributed to these methods are not designed for long context alignment and understanding, thus failing to select the samples enriched with long-range dependency relations. 
We observe that using 30\% of the entire long instruction-following dataset (\textit{GATEAU-LLaMA-30\%}) gives the best LongBench performance in both settings. This is likely because it strikes a good balance between the amount and quality of long instruction-following samples, leading to better results.

\subsection{Analysis}

\begin{table*}[t]
\centering  
\scriptsize
\resizebox{\textwidth}{!}{
\begin{tabular}{lccccccccccccccc}
\toprule
\multirow{2}{*}{\textbf{Model}} & \multicolumn{3}{c|}{\textbf{LongBench}} & \multicolumn{1}{c|}{\textbf{LongBench-Chat}} & \multicolumn{3}{c}{\textbf{MT-Bench}} \\
\cmidrule(lr){2-4} \cmidrule(lr){5-5} \cmidrule(lr){6-9} 
& \textbf{Single-Doc QA} & \textbf{Multi-Doc QA} & \textbf{Summarization} & \textbf{Avg} & \textbf{First-turn} & \textbf{Second-turn} & \textbf{Avg} 
\\
\midrule
\multicolumn{16}{c}{\cellcolor{mypink}\textbf{
Real-world Settings}} \\
\rowcolor{blue!5} \textbf{GATEAU-LLaMA - 13B - 50\%} &\textbf{40.2}&\textbf{27.1}&\textbf{25.7}&\textbf{61.4} &\textbf{66.8}&\textbf{55.3}&\textbf{61.1}   \\
\ -w/o Contextual Awareness Measurement & 38.1  & 25.8 & 24.6 & 60.2 & 66.2 &	55.0 &	60.6 \\
\ -w/o Homologous Models’ Guidance & 38.6  & 26.0 & 25.1 & 60.6 & 66.0 &	54.6 &	60.3 \\
\ -w/o Data Selection (i.e., Full - 100\%) & 33.6 & 16.7 & 24.4 & 59.4 & 66.0& 	54.1 	&59.6 \\
\rowcolor{blue!5} \textbf{GATEAU-LLaMA - 7B - 50\%} & \textbf{38.9}   & \textbf{25.8}   & \textbf{25.5}   & \textbf{56.8} & \textbf{64.1}	& \textbf{50.4}	& \textbf{57.3} \\
\ -w/o Contextual Awareness Measurement & 38.4   & 24.3  & 25.1  & 53.2
& 61.7	&51.5	&56.6\\
\ -w/o Homologous Models’ Guidance & 38.6  & 24.5  & 24.9  & 52.8& 63.1	&49.3	&56.3\\
\ -w/o Data Selection (i.e., Full - 100\%) & 36.1  & 22.3   & 23.8   &48.8&
60.0	&48.7&	54.3\\

\midrule
\multicolumn{16}{c}{\cellcolor{mypink}\textbf{
Limited Short Instruction Data Settings}} \\
\rowcolor{blue!5} \textbf{GATEAU-LLaMA - 13B - 50\%} & \textbf{32.1} & \textbf{19.1} & \textbf{25.3} & \textbf{62.6} & \textbf{66.0} 	& \textbf{51.5} &	\textbf{58.8} \\
\ -w/o Contextual Awareness Measurement & 31.4  & 18.4 & 24.7 & 59.6 & 64.2 &	50.3 &	57.3 \\
\ -w/o Homologous Models’ Guidance & 30.8  &18.6 & 25.0 & 60.4 & 63.6 & 50.6 &	57.1 \\
\ -w/o Data Selection (i.e., Full - 100\%) &30.4 & 17.8  & 24.5 & 54.2 & 61.0	& 49.8 &	55.4\\
\rowcolor{blue!5} \textbf{GATEAU-LLaMA - 7B - 50\%} & \textbf{31.0}  & \textbf{18.1}  & \textbf{25.3}  & \textbf{59.0} & \textbf{64.2} &	\textbf{44.1} &	\textbf{54.2}  \\
\ -w/o Contextual Awareness Measurement & 28.5 & 17.5  &24.7  &53.2&
61.3 &	42.4 &	51.8 
		\\
\ -w/o Homologous Models’ Guidance
 & 28.7  &17.3 & 24.6 & 54.4 &
56.1 &	45.0 &	50.6 
		\\
\ -w/o Data Selection (i.e., Full - 100\%) &27.2 & 16.1 & 24.5 & 50.8&
54.5 &	40.9 	&47.7 
		\\

\bottomrule
\end{tabular}
}
\caption{Results (\%) of ablation and scalability study. We show automatic metrics evaluation results on LongBench. 
}
\label{tb:abs_sca}
\end{table*}

\begin{figure}
    \centering
    \includegraphics[width=0.85\linewidth]
    {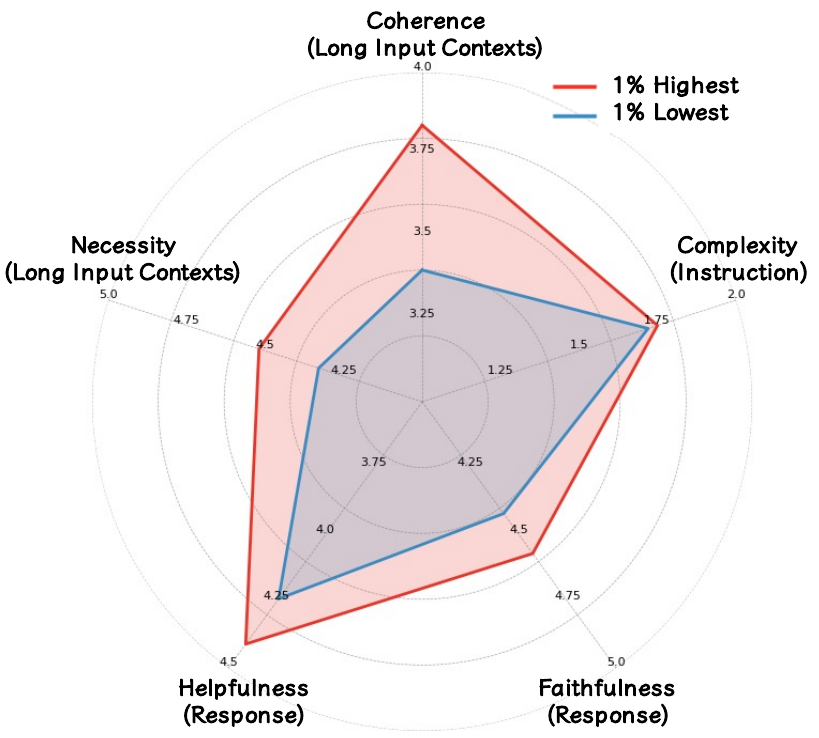}
    \caption{
    The comparison between samples with top 1\% and least 1\% scored by GATEAU.}
    \label{fig_charactercteristics}
\end{figure}


\noindent
\textbf{Ablation Study.}
To evaluate the effectiveness of our proposed GATEAU, we also conduct the ablation study in Table \ref{tb:abs_sca}.
We can find that HMG and CAM can both enhance LLMs’ instruction-following and long-context understanding capabilities.
This indicates the effectiveness of GATEAU, and using the two proposed methods can further improve the performance as they separately measure the difficulty from two different perspectives.

\noindent
\textbf{Scalability Study.}
We explore whether GATEAU can fit in larger LLMs in Table \ref{tb:abs_sca}.
We apply our method on the Llama-2-13B-base series and fine-tune Llama-2-13B-base-64k \citep{bai2024longalignrecipelongcontext} using the selected samples.
Compared to the 7B-scale \textit{GATEAU-LLaMA-7B}, the 13B \textit{GATEAU-LLaMA-13B} shows consistent improvements on three benchmarks.
This indicates that GATEAU scales effectively to larger-scale models.

\noindent
\textbf{General Characteristics of Selected Samples.}
We delve into whether the selected samples based on GATEAU align with known characteristics of high-quality data as shown in Figure \ref{fig_charactercteristics}.
We select 100 samples with the 1\% highest scores and 100 samples with the 1\% lowest scores.
Utilizing GPT-4, we evaluate each sample on five aspects: the coherence of long input contexts, the necessity of long input contexts, helpfulness of response, the faithfulness of response, and the complexity of instruction.
A sample with a higher score tends to be more high-quality, especially the long input contexts and the response.
The complexity of instruction, in particular, shows a mere improvement compared to other characteristics.
We evaluate the whole dataset on this characteristic and find that all samples show consistently low scores, which may be due to the limitation of the synthetic dataset.
More details are shown in the Appendix \ref{char_section}. 

\begin{figure*}
    \centering
    \includegraphics[width=0.5\textheight]{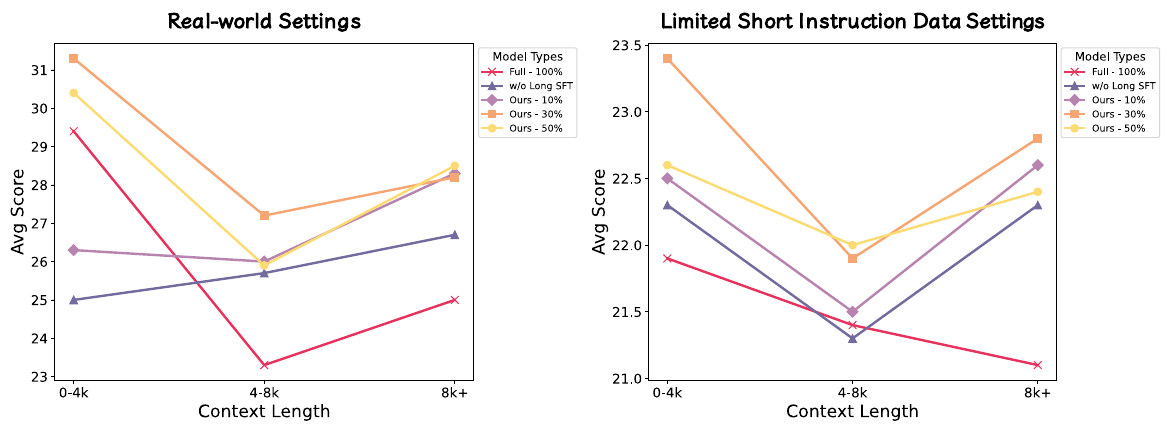}
    \caption{
    Automatic metrics evaluation results (\%) under different context lengths on LongBench.
    }
    \label{fig_length}
\end{figure*}

\begin{figure}[t]
    \hspace{-10mm}
    \centering
    \includegraphics[width=0.27\textheight]
    {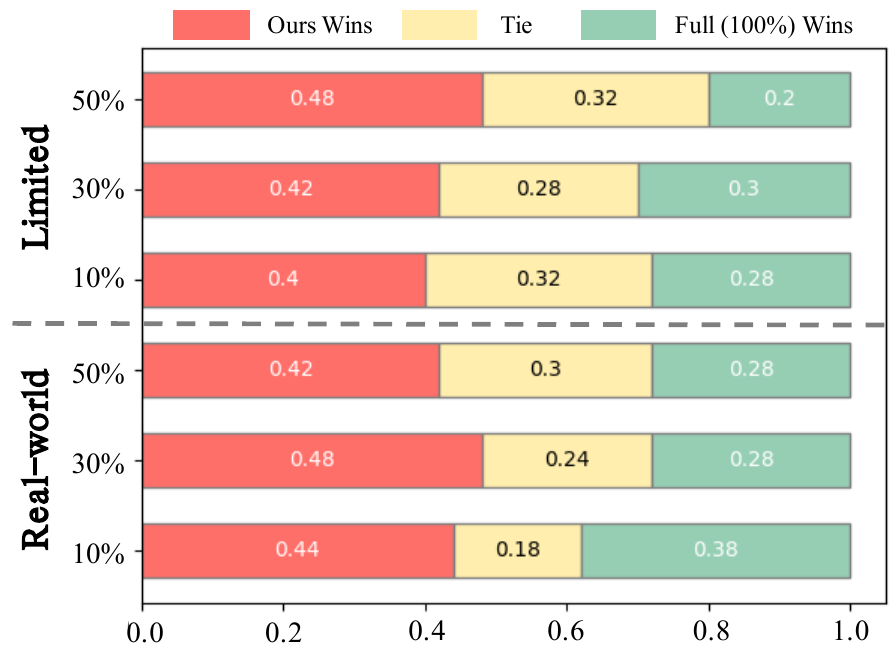}
    \caption{
    Human evaluation in two settings.}
    \label{fig_human}
\end{figure}

\noindent
\textbf{Variation of Abilities under Different Context Lengths.}
Figure \ref{fig_length} shows the macro-average results (\%) on data in length ranges of 0-4k, 4k-8k, and 8k+.
We find that GATEAU improves the performance in long-context scenarios (i.e., 4k-8k and 8k+).
Indiscriminately using the full long SFT dataset (\textit{Full-100\%}) even hinders the performance in long-context scenarios compared to solely using the short instruction-following dataset (\textit{-w/o Long SFT}).
This confirms the necessity of selecting influential samples and the effectiveness of GATEAU.

\noindent
\textbf{Human Evaluation.}
We conduct a human evaluation on the LongBench-Chat.
We invite three participants (Ph.D. students or Master students) to compare the responses generated by the models.
For each comparison, three options are given (Win, Tie, and Loss), and the final result is determined by majority voting.
Figure \ref{fig_human} shows the effectiveness of our method.
The participants follow the principles in Figure~\ref{fig:human_evaluation_principles} to make the decision.
If the final result can not be determined by majority voting, we will hold a discussion among the participants and vote on the result again. 

\noindent
\textbf{Needle in the Haystack Test.}
We conduct ``Needle in A HayStack'' experiments in Figure \ref{fig_neddle} to test the model’s ability to utilize information from 10 different positions. 
This task asks for the model to retrieve a piece of fact (the ‘needle’) that is inserted in the middle (positioned at a specified depth percent) of a long context window (the ‘haystack’).
These results show that GATEAU can help LLMs to utilize information from different positions within long texts, resulting in a decrease in the model’s retrieval error.
Following the same original configuration as the original method \citep{needlehay}, we use ``The best thing to do in San Francisco is eat a sandwich and sit in Dolores Park on
a sunny day.''  as the needle fact, and Paul Graham’s essays as the long haystack context.
We use the same prompt as \citet{bai2024longalignrecipelongcontext} to test our method.



\noindent
\textbf{Parameter Study and Case Study.}
\
We also conduct the parameter study and a practical case study in the Appendix \ref{appendix:parameter} and Appendix \ref{appendix:case}.

\noindent
\textbf{Discussion.}
\
We further discuss some possible concerns about GATEAU in the Appendix \ref{appendix:dis}.
For example, we report the execution time of GATEAU.

%% file: Files/2_Related_work.tex
\section{Related Work}

\textbf{Long Context Alignment.}
Aligning the LLMs to follow user instructions has been extensively studied in short-context scenarios \citep{wang-etal-2023-self-instruct, an2025ultraifadvancinginstructionfollowing, si2025teachinglargelanguagemodels}.
However, excessively long contexts present unique challenges for long context alignment.
\citet{longchat2023,tworkowski2023focused,xiong-etal-2024-effective} construct the long SFT dataset by concatenating short SFT samples.
Yet, simply concatenating unrelated sentences can not effectively simulate the long-range dependency relations for long-context tasks.
Thus, \citet{yang2023longqlora,chen2024longlora, bai2024longalignrecipelongcontext} construct long SFT data by collecting long-context materials as inputs and querying Claude to get the response.
However, using these synthetic data without a clear strategy for ensuring data quality may introduce low-quality samples (e.g., samples without meaningful long-range dependency relations).
Training LLMs on such low-quality samples can constrain their final performance.

\noindent
\textbf{Data Selection for Alignment.}
As \citet{zhou2023lima} states \textit{less is more for alignment}, many works attempt to select influential samples to empower the LLMs’ instruction-following capabilities.
\citet{chen2023alpagasus, liu2024what} attempt to utilize the feedback from well-aligned closed-source LLMs to select samples.
\citet{cao2024instructionmininginstructiondata,li-etal-2024-quantity, ge2024clustering, xia2024less, si-etal-2025-aligning} try to utilize the well-designed metrics (e.g., complexity) based on open-source LLMs to rank and select the samples.
Meanwhile, \citet{li2024shot,zhang2024recostexternalknowledgeguided} attempt to utilize the guidance from in-context learning.
However, these methods only focus on selecting short SFT data, ignoring the unique challenge in long context alignment, i.e., selecting the samples enriched with meaningful long-range dependency relations.

%% file: Files/5_Conclusion.tex
\section{Conclusion}
\label{subsection:conclusion}
In this study, we introduce \textbf{GATEAU}, a new novel framework designed to select influential samples for long context alignment.
Different from previous studies, we attempt to address the unique challenge in long context alignment, i.e., modeling long-range dependencies. 
To measure the richness of long-range dependency relations in long SFT samples, GATEAU separately measures the difficulty of generating corresponding responses and understanding lengthy inputs due to the long-range dependencies.
Trained on these selected influential samples, our model achieves better alignment. 
Extensive experiments consistently show the effectiveness of GATEAU compared to other methods.

\begin{figure*}[t]
    \centering
    \includegraphics[scale=0.2]{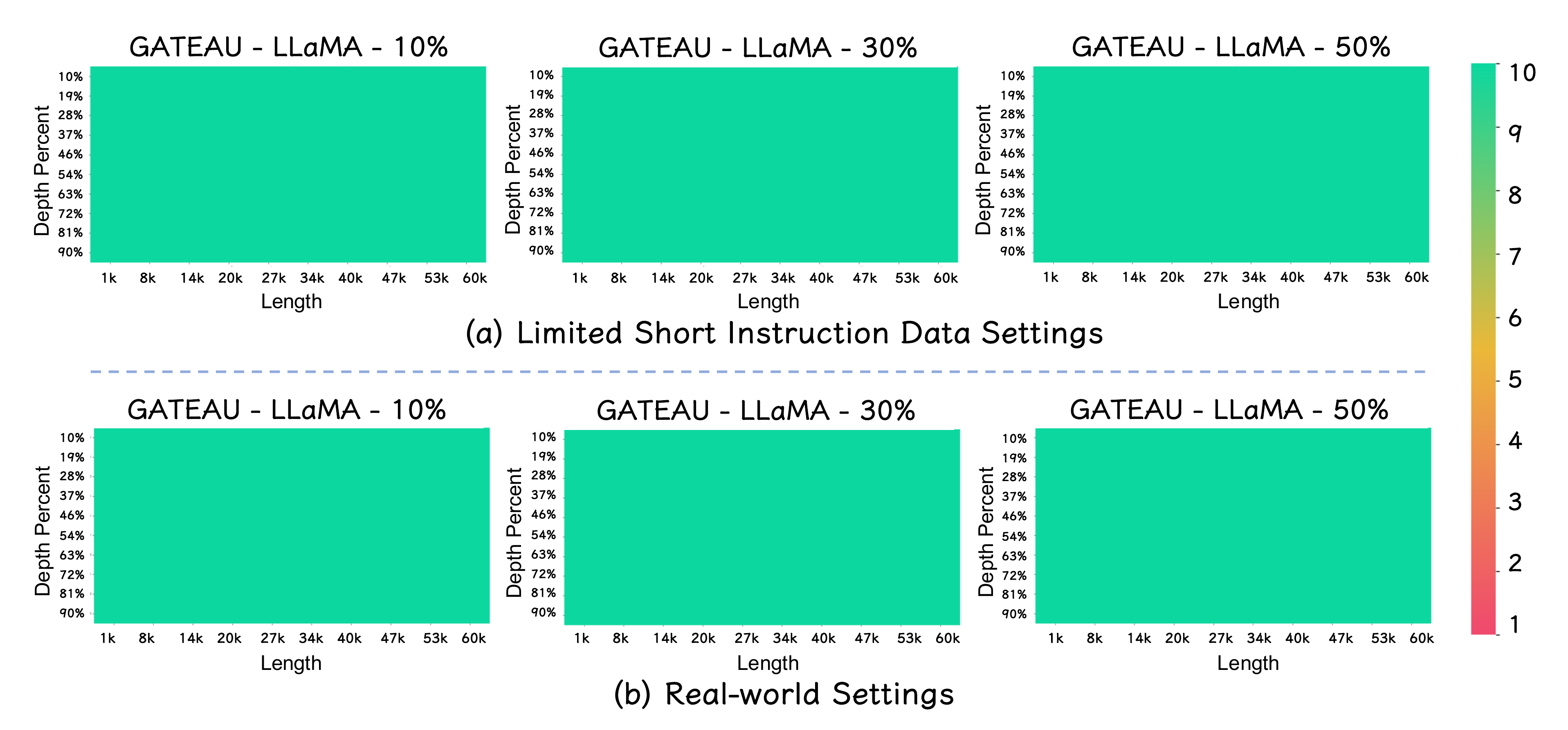}
    \caption{
    Needle in the Haystack test.
    }
    \label{fig_neddle}
\end{figure*}

\section*{Limitations}
Although empirical experiments have confirmed the efficacy of the proposed GATEAU, three major limitations remain.
Firstly, our proposed HMG requires two homologous models with different context windows, thus limiting the range of models we can use to conduct more experiments in our paper.
However, in practical scenarios, training a powerful long-context LLM always involves homologous models with different context windows (though these models may not be open-sourced). 
This is because existing LLMs are often initially pre-trained on a large-scale corpus with smaller context windows due to device limitations, they then conduct continual pre-training to extend the window size. Therefore, our method still remains effective in real-world scenarios.
Secondly, GATEAU is designed to improve overall performance in instruction-following and long-context understanding tasks. 
It is not suitable to improve the performance of LLMs in a targeted capability or task, e.g., mathematical questions.
The size of the context window is a crucial consideration, typically determined during the continual training phase of open-source foundation models. 
When input data exceeds the context window of the underlying model, the performance of GATEAU may be constrained. 
Nevertheless, ongoing advancements in open-source models are driving significant increases in context window sizes \citep{yang2025qwen251mtechnicalreport}. 
Consequently, leveraging open-source long-context large language models, GATEAU can be further employed to select and process longer samples.

\section*{Acknowledgements}
We would like to thank the anonymous reviewers for their suggestions.
This work is supported by the National Natural Science Foundation of China (No. 62236011, No.T2341003) and a grant from the Guoqiang Institute, Tsinghua University.

%% file: Files/6_Appendix.tex
\appendix

\section*{Appendix}
\noindent This appendix is organized as follows.  

\begin{itemize}  
    \item In Section~\ref{appendix:training}, we report the training details, e.g., training datasets and hyperparameters.  
    \item  In Section~\ref{appendix:baselines}, we go into detail about the baselines used in our experiments.
    \item In Section~\ref{appendix:evaluations}, we show the details of evaluations, e.g., the introduction of the used benchmarks and evaluation prompts.
    \item In Section~\ref{char_section}, we list the details of the general characteristics of selected samples.
    \item In Section~\ref{appendix:parameter}, we conduct experiments to explore the impact of hyperparameters.
    \item In Section~\ref{appendix:case}, we come up with a practical case study to show the effectiveness of GATEAU.
    \item In Section~\ref{appendix:dis}, we discuss some possible questions, including execution time (Sec. \ref{appendix:dis_time}), experiments in other LLMs (Sec. \ref{appendix:dis_other}), 
    experiments in other long SFT datasets (Sec. \ref{appendix:dis_other_datasets}), the diversity of selected samples (Sec. \ref{appendix:dis_diversity}), further exploration of HMG (Sec. \ref{appendix:dis_hmg}), orthogonality with baselines (Sec. \ref{appendix:dis:orth}), and mix-quality training strategy (Sec. \ref{appendix:dis-mix}).
    \item In Section~\ref{appendix:csc}, we show the difference between samples with high or low scores.

\end{itemize}

\section{Training}
\label{appendix:training}

\textbf{Training Datasets.}
\
LongAlign dataset \citep{bai2024longalignrecipelongcontext} is developed by using collected long sequences from 9 sources and applying the Self-Instruct \citep{wang-etal-2023-self-instruct} approach with long-context LLM Claude 2.1 \citep{claude-2-1-2023}.
Though initially competitive, its dependence on Claude 2.1 synthesized data may lead to quality concerns.
For the ShareGPT dataset \citep{sharegpt}, we filter the sample with an empty response.

\noindent
\textbf{Training Details.}
\
LLaMA-2-7B-base-4k is an open-sourced LLM with a context window of 4k tokens.
To extend context windows, \citet{bai2024longalignrecipelongcontext} proposes LLaMA-2-7B-base-64k by 
modifying the RoPE position encoding \citep{su2023roformerenhancedtransformerrotary} and 
applying continual training on data with lengths under 64k, for a total of 10 billion tokens.
Meanwhile, for LLaMA-2-7B-base-4k, we expand the base frequency $b$ of the RoPE position encoding by 200 times (from 10,000 to 2,000,000) to extend the context windows and avoid the model conducting extreme perplexity score (>1,000) in HMG.
For CAM, we use LLaMA-2-7B-base-64k to calculate the score and use selected samples to train the LLaMA-2-7B-base-64k as our final model.

\noindent
\textbf{Devices and Hyperparameters.}
\
All models are trained with 8xA800 80G GPUs and DeepSpeed+ZeRO3+CPU offloading.
We use BF16 in both our training and inference.
The models can be trained with a maximum length of 64k tokens without GPU memory overflow. 
We set the maximum length of the training data to 64k, with any data exceeding this length being truncated from the right side. 
We keep the same maximum length in the HMG and CAM, but truncate from the left side to keep the original responses.
We set the batch size to 8, with a gradient accumulation step of 12 for all the training methods.
We train 2 epochs on the training data.
We set the learning rate as 2e-5 and use AdamW \citep{loshchilov2018decoupled} as our optimizer.
The $\beta_1$ and $\beta_2$ in the AdamW optimizer are set to 0.9 and 0.95.
Meanwhile, the length of segment $L$ is set to 128 in CAM.
Hyperparameter $\alpha$ in Eq. (\ref{final_score}) is set to 0.7 in Limited Short Instruction Data Settings and 0.8 in Real-world Settings.

\section{Baselines}
\label{appendix:baselines}
We will detail the baselines in our experiments.

\noindent
\textbf{w/o SFT.}
\
For w/o SFT, we directly utilize the base model without alignment to get the experiment results, i.e., the results of LLaMA-2-7B-base-64k.

\noindent
\textbf{w/o Long SFT.}
\
For baseline w/o Long SFT, we only use the short instruction data from the ShareGPT dataset to apply the supervised fine-tuning stage for alignment.
The number of short instruction samples used from the ShareGPT dataset is determined by the different settings.

\noindent
\textbf{Full - 100\%.}
\
For baseline Full - 100\%, we use the full data of the LongAlign dataset, including 10k long instruction samples, to conduct the SFT for alignment.
The number of short instruction samples used from the ShareGPT dataset is determined by the different settings.

\noindent
\textbf{Perplexity Guidance.}
\
We use the perplexity score from LLM as guidance to select long instruction-following samples according to Eq. (\ref{equal-ppl}).
We select the long samples with the highest perplexity scores to train the model.
The number of short instruction samples used from ShareGPT is determined by the different settings.

\begin{table*}[t]
\centering  
\scriptsize
\resizebox{0.95\textwidth}{!}{
\begin{tabular}{lclrccc}
\toprule
Dataset & ID & Source & Avg len & Auto Metric & Language & \#data \\
\midrule
\cellcolor{myyellow} \emph{Single-Document QA} \\
NarrativeQA & 1-1 & Literature, Film & 18,409 & F1 & English & 200 \\
Qasper & 1-2 & Science & 3,619 & F1 & English & 200 \\
MultiFieldQA-en & 1-3 & Multi-field & 4,559 & F1 & English & 150 \\
 MultiFieldQA-zh &  1-4 &  Multi-field &  6,701 &  F1 &  Chinese &  200 \\
\midrule
\cellcolor{myyellow} \emph{Multi-Document QA} \\
HotpotQA & 2-1 & Wikipedia & 9,151 & F1 & English & 200 \\
2WikiMultihopQA & 2-2 & Wikipedia & 4,887 & F1 & English & 200 \\
MuSiQue & 2-3 & Wikipedia & 11,214 & F1 & English & 200 \\
 DuReader &  2-4 &  Baidu Search &  15,768 &  Rouge-L &  Chinese &  200 \\
\midrule
\cellcolor{myyellow} \emph{Summarization} \\
GovReport & 3-1 & Government report & 8,734 & Rouge-L & English & 200 \\
QMSum & 3-2 & Meeting & 10,614 & Rouge-L & English & 200 \\
MultiNews & 3-3 & News & 2,113 & Rouge-L & English & 200 \\
 VCSUM &  3-4 &  Meeting &  15,380 &  Rouge-L &  Chinese &  200 \\

\bottomrule
\end{tabular}
}
\caption{An overview of the dataset statistics in LongBench. `Source' denotes the origin of the context. `Avg len' is computed using the number of words for the English datasets and the number of characters for the Chinese datasets.}
\label{table:longbench-stat}
\end{table*}

\noindent
\textbf{CaR.}
\
This work \citep{ge2024clustering} proposes a straightforward yet efficacious short instruction-following selection framework.
This method first selects a subset that ensures the retention of a large number of high-quality instructions and then supplements a small number of high-quality instructions from each cluster to enhance the diversity of the data while preserving instruction quality.
Specifically, this work first employs a small-scale trained reward model to get the score of the samples.
Meanwhile, the cluster model is employed to cluster all candidate instruction pairs into $k$ clusters
Finally, all instruction pairs are sorted based on their scores, and the top $n_1$ pairs are selected; within each cluster, instruction pairs are sorted by score, and the top $n_2$ pairs are chosen.
A high-quality sub-dataset with preserved diversity is then curated by duplicating $n_1 + k \times n_2$ pairs of instructions.
We directly use the same reward model and hyperparameters to select long samples.
The number of short samples used from ShareGPT is determined by the different settings. 

\noindent
\textbf{Cherry Selection.}
\
\citet{li-etal-2024-quantity} proposes a method for sifting through expansive open-source short instruction-following datasets to discover the most influential training samples.
This work introduces a new metric called the Instruction-Following Difficulty (IFD) score, which compares how well models respond to an instruction versus how well they respond without one.
This method measures how difficult each sample is for the model and uses samples with high IFD scores to further train the model.
We apply this method to select the long samples as the baseline.
The number of short instruction samples used from ShareGPT is determined by the different settings.

\begin{table*}[!h]
\scriptsize
\centering  
\resizebox{\textwidth}{!}{
\begin{tabular}{lccccccccccccccc}
\toprule
\multirow{2}{*}{\textbf{Model}} & \multicolumn{5}{c}{\textbf{Single-Doc QA}} & \multicolumn{5}{c}{\textbf{Multi-Doc QA}} & \multicolumn{5}{c}{\textbf{Summarization}} \\
\cmidrule(lr){2-6} \cmidrule(lr){7-11} \cmidrule(lr){12-16} 
& \textbf{1-1} & \textbf{1-2} & \textbf{1-3} & \textbf{1-4} & \textbf{\textcolor[rgb]{0.8,0,0}{Avg}} & \textbf{2-1} & \textbf{2-2} & \textbf{2-3} & \textbf{2-4} & \textbf{\textcolor[rgb]{0.8,0,0}{Avg}} & \textbf{3-1} & \textbf{3-2} & \textbf{3-3} & \textbf{3-4} & \textbf{\textcolor[rgb]{0.8,0,0}{Avg}} \\
\midrule
\rowcolor{mygray} \multicolumn{16}{c}{\cellcolor{mypink} \textbf{Real-world Settings}} \\
w/o SFT & 0.9 	&3.9 	&6.4 & 3.6	&3.7 	&7.3  &8.7	&2.1&15.4 	&8.4 	&23.9 &	6.2 	& 14.0 &1.8 	&11.5  \\
w/o Long SFT	&16.8 	&29.1 	&45.8	&48.7	&35.1 	&27.8	&17.6	&11.4	&25.3	&20.5 	&27.4 	&23.3 	&27.8 	&14.3 	&23.2 \\
Full - 100\% & 18.4	&29.9	&46.1	&\textbf{49.9} &	36.1 &	27.1 	&20.8 	&11.2 	&30.0 &	22.3 	&28.7 	&24.0 &	26.7 	&15.9 &	23.8 \\
\midrule
Perplexity Guidance - 10\%	&19.9 &	32.0 	&46.6 &	45.8 	&36.1 &	22.1 	&23.2 &	10.4 	&30.3 &	21.5 	&31.3 	&23.8 	&26.0 &	17.7 &	24.7 \\
CaR -  10\%	&16.9	&24.1	&47.6	&42.3	&32.7 	&22.1	&19.8	&11.3	&30.0	&20.8	&\textbf{31.9}	&23.1	&26.2	&18.6	&25.0 \\
Cherry Selection - 10\%	&19.9 	&30.8 	&47.2 &	43.1 &	35.3 	&25.2 &	21.4 	&10.6 &	28.3 	&21.4 &	30.0 &	24.1 &	25.1 	&17.0 &	24.1 \\
\rowcolor{blue!5} \textbf{GATEAU-LLaMA - 10\%} &\textbf{23.5} 	&\textbf{34.2} 	&\textbf{49.6} 	&\textbf{54.5} 	&\textbf{40.5} 	&\textbf{28.7} 	&\textbf{25.0} 	&\textbf{12.1} 	&\textbf{30.5} 	&\textbf{24.0} 	&31.2 	&\textbf{24.7} 	&\textbf{26.9} 	&\textbf{18.9} 	&\textbf{25.4}  \\ 
\rowcolor{blue!5}  $\Delta$ compared to Full - 100\%  
&  
\textcolor[rgb]{0.7,0,0}{+5.1} 	&  
\textcolor[rgb]{0.7,0,0}{+4.3} 	&  
\textcolor[rgb]{0.7,0,0}{+3.5} 	&  
\textcolor[rgb]{0.7,0,0}{+4.6} 	&  
\textcolor[rgb]{0.7,0,0}{+4.4} 	&  
\textcolor[rgb]{0.7,0,0}{+1.6} 	&  
\textcolor[rgb]{0.7,0,0}{+4.2} 	&  
\textcolor[rgb]{0.7,0,0}{+0.9} 	&  
\textcolor[rgb]{0.7,0,0}{+0.5} 	&  
\textcolor[rgb]{0.7,0,0}{+1.8} 	&  
\textcolor[rgb]{0.7,0,0}{+2.5} 	&  
\textcolor[rgb]{0.7,0,0}{+0.7} 	&  
\textcolor[rgb]{0.7,0,0}{+0.2} 	&  
\textcolor[rgb]{0.7,0,0}{+3.0}	&  
\textcolor[rgb]{0.7,0,0}{+1.6}  \\ 

\midrule
Perplexity Guidance -  30\%	&21.1 	&33.6 &	46.1 	&46.7 &	36.9 &	23.4 	&21.0 	&10.1 	&30.1& 	21.2 &	30.2	&24.7&	26.4&	18.9	&25.1   \\
CaR -  30\%	&18.0	&24.4	&46.9	&45.0	&33.6 	&25.4	&20.8	&14.4	&29.4	&22.5	&30.1	&24.8	&26.5	&18.2	&24.9\\
Cherry Selection - 30\%	&20.5 	&33.1 	&48.0 	&51.0 	&38.2 	&26.7 	&20.4 	&13.5 	&29.1 	&22.4 	&30.4 	&24.1 	&26.9 	&17.7 	&24.8 \\
\rowcolor{blue!5} \textbf{GATEAU-LLaMA - 30\%}	&\textbf{23.7} 	&\textbf{34.1} 	&\textbf{49.6} 	&\textbf{54.6} 	&\textbf{40.5} 	&\textbf{30.1} 	&\textbf{23.8} 	&\textbf{14.9} 	&\textbf{30.4} 	&\textbf{24.8} 	&\textbf{30.5}	&\textbf{24.9} 	&\textbf{27.2} 	&\textbf{18.9} 	&\textbf{25.4}   \\
\rowcolor{blue!5}  $\Delta$ compared to Full - 100\%  &  \textcolor[rgb]{0.7,0,0}{+5.3} 	&  \textcolor[rgb]{0.7,0,0}{+4.2} 	&  \textcolor[rgb]{0.7,0,0}{+3.5} 	&  \textcolor[rgb]{0.7,0,0}{+4.7} 	&  \textcolor[rgb]{0.7,0,0}{+4.4} 	&  \textcolor[rgb]{0.7,0,0}{+3.0} 	&  \textcolor[rgb]{0.7,0,0}{+3.0} 	&  \textcolor[rgb]{0.7,0,0}{+3.7} 	&  \textcolor[rgb]{0.7,0,0}{+0.4} 	&  \textcolor[rgb]{0.7,0,0}{+2.5} 	&  \textcolor[rgb]{0.7,0,0}{+1.8} 	&  \textcolor[rgb]{0.7,0,0}{+0.9} 	&  \textcolor[rgb]{0.7,0,0}{+0.5} 	&  \textcolor[rgb]{0.7,0,0}{+3.0} 	&  \textcolor[rgb]{0.7,0,0}{+1.6} \\

\midrule
Perplexity Guidance - 50\%	&19.2 	&32.8 &	50.1 &	49.5 &	37.9  	&27.1 	&23.1 	&12.1 	&31.1 	&23.4 	&\textbf{31.5} 	&24.1 	&27.1 	&18.7 	&25.4 \\
CaR -  50\%	&17.6 	&24.5 	&47.6 	&44.7 	&33.6 	&29.3 	&19.4 	&17.3 	&29.6 	&23.9 	&30.3 	&23.7 	&26.0 	&18.2 	&24.6  \\
Cherry Selection - 50\%	&19.0 &	32.6 	&51.7 &	49.6 &	38.2  	&26.2 	&23.9 	&13.5 	&30.4 	&23.5 	&30.5 	&23.8 	&26.9 	&18.8 	&25.0 \\
\rowcolor{blue!5} \textbf{GATEAU-LLaMA - 50\%}	&
\textbf{20.2} 	&
\textbf{33.4} 	&
\textbf{52.1} 	&
49.8 	&
\textbf{38.9} 	&
\textbf{30.7} 	&
\textbf{25.2} 	&
\textbf{15.0} 	&
\textbf{32.5} 	&
\textbf{25.8} 	&
31.3 	&
\textbf{24.6} 	&
\textbf{27.1} 	&
\textbf{18.8} 	&
\textbf{25.5}  \\
\rowcolor{blue!5}  $\Delta$ compared to Full - 100\%  &  \textcolor[rgb]{0.7,0,0}{+1.8} 	&  \textcolor[rgb]{0.7,0,0}{+3.5} 	&  \textcolor[rgb]{0.7,0,0}{+6.0} 	&  \textcolor[rgb]{0,0.7,0}{-0.1}	&  \textcolor[rgb]{0.7,0,0}{+2.8} 	&  \textcolor[rgb]{0.7,0,0}{+3.6} 	&  \textcolor[rgb]{0.7,0,0}{+4.4} 	&  \textcolor[rgb]{0.7,0,0}{+3.8} 	&  \textcolor[rgb]{0.7,0,0}{+2.5} 	&  \textcolor[rgb]{0.7,0,0}{+3.6} 	&  \textcolor[rgb]{0.7,0,0}{+2.6} 	&  \textcolor[rgb]{0.7,0,0}{+0.6} 	&  \textcolor[rgb]{0.7,0,0}{+0.4} 	&  \textcolor[rgb]{0.7,0,0}{+2.9} 	&  \textcolor[rgb]{0.7,0,0}{+1.6}  \\
\midrule
\rowcolor{mygray} \multicolumn{16}{c}{\cellcolor{mypink} \textbf{
Limited Short Instruction Data Settings}} \\
w/o SFT & 0.9	&3.9	&6.4	&3.6	&3.7 	&7.3&	8.71	&2.1	&15.4	&8.4	&23.9&	6.2	&14.0	&1.78	&11.5 \\
w/o Long SFT &  13.8 &	19.2 &	38.3 &	37.1 	&27.1 	&15.2	&14.7	&8.2&	25.7	&16.0 &	29.4 &	24.4 &	25.0 	&19.3 &	24.5 \\
Full - 100\% & 14.7 &	20.1 &	37.0 	&37.0& 	27.2 &	15.4 &	13.8 &	8.6 &	26.7 	&16.1 &	29.3 	&24.5 &	25.6 	&18.6& 	24.5 \\
\midrule
Perplexity Guidance - 10\%	&15.4 &	19.2 &	41.0 &	37.8 &	28.4 &	15.0 &	14.8 &	\textbf{8.5} 	&25.6 &	16.0 &	28.8  &	23.9  &	26.1 & 	17.8 	 &24.2   \\
CaR -  10\%	&11.5 &	17.7 &	37.7 &	30.0 &	24.2 	&15.6 &	12.5 &	8.4 	&25.9 &	15.6  &	29.3  &	24.1  &	26.2  &	18.2 	 &24.5  \\
Cherry Selection - 10\%	&14.6 & 	19.2 &	41.2 &	37.7 &	28.2 &	15.7 	&14.6 &	7.6 	&25.3 &	15.8 &	29.4 &	24.1 	&26.0 &	17.8 &	24.3  \\
\rowcolor{blue!5} \textbf{GATEAU-LLaMA - 10\%}   & \textbf{17.1} 	& \textbf{20.7} 	& \textbf{43.4} 	& \textbf{38.3} 	& \textbf{29.9} 	& \textbf{19.9} 	& \textbf{18.5} 	& 8.2 	& \textbf{26.8} 	& \textbf{18.4} 	& \textbf{29.6} 	& \textbf{24.3} 	& \textbf{26.3} 	& \textbf{18.3} 	& \textbf{24.6} \\
\rowcolor{blue!5}  $\Delta$ compared to Full - 100\%  &  \textcolor[rgb]{0.7,0,0}{+2.4} 	&  \textcolor[rgb]{0.7,0,0}{+0.6} 	&  \textcolor[rgb]{0.7,0,0}{+6.4} 	&  \textcolor[rgb]{0.7,0,0}{+1.3} 	&  \textcolor[rgb]{0.7,0,0}{+2.7} 	&  \textcolor[rgb]{0.7,0,0}{+4.5} 	&  \textcolor[rgb]{0.7,0,0}{+4.7} 	&  \textcolor[rgb]{0,0.7,0}{-0.4}	&  \textcolor[rgb]{0.7,0,0}{+0.1} 	&  \textcolor[rgb]{0.7,0,0}{+2.2} 	&  \textcolor[rgb]{0.7,0,0}{+0.3} 	&  \textcolor[rgb]{0,0.7,0}{-0.2}	&  \textcolor[rgb]{0.7,0,0}{+0.7} 	&  \textcolor[rgb]{0,0.7,0}{-0.3}	&  \textcolor[rgb]{0.7,0,0}{+0.1}   \\

\midrule
Perplexity Guidance -  30\%	&15.3 &	\textbf{20.6} &	42.3 &	38.2 &	29.1 &	17.4& 	15.9 &	8.6 &	27.5 	&17.4  &	28.3 &	24.3 &	25.7 	&19.0 	&24.3  \\
CaR -  30\%	&13.6 &	18.3 &	41.0 &	30.5 &	25.9 	&16.7 &	15.8 &	9.4 &	27.0 &	17.2  &	28.8 &	24.3 	&25.3& 	18.4 &	24.2 \\
Cherry Selection - 30\%	&15.9 &	19.5 &	42.3 &	\textbf{39.0} &	29.2  &	17.3 &	16.3 &	9.3 	&26.2& 	17.3 	&29.2 &	\textbf{25.0} &	26.1 &	18.2 &	24.6  \\
\rowcolor{blue!5} \textbf{GATEAU-LLaMA - 30\%} &\textbf{17.7} 	&20.4 	&\textbf{43.1} 	&38.6&\textbf{29.9} 	&\textbf{22.5} 	&\textbf{18.5} 	&\textbf{11.6} 	&\textbf{27.7} 	&\textbf{20.1} 	&\textbf{30.5} 	&24.3 	&\textbf{26.8} 	&\textbf{19.7} 	&\textbf{25.3}  \\

\rowcolor{blue!5}  $\Delta$ compared to Full - 100\%  &  \textcolor[rgb]{0.7,0,0}{+3.0} 	&  \textcolor[rgb]{0.7,0,0}{+0.3} 	&  \textcolor[rgb]{0.7,0,0}{+6.1} 	&  \textcolor[rgb]{0.7,0,0}{+1.6} 	&  \textcolor[rgb]{0.7,0,0}{+2.7} 	&  \textcolor[rgb]{0.7,0,0}{+7.1} 	&  \textcolor[rgb]{0.7,0,0}{+4.7} 	&  \textcolor[rgb]{0.7,0,0}{+3.0} 	&  \textcolor[rgb]{0.7,0,0}{+1.0} 	&  \textcolor[rgb]{0.7,0,0}{+4.0} 	&  \textcolor[rgb]{0.7,0,0}{+1.2} 	&  \textcolor[rgb]{0,0.7,0}{-0.2}	&  \textcolor[rgb]{0.7,0,0}{+1.2} 	&  \textcolor[rgb]{0.7,0,0}{+1.1} 	&  \textcolor[rgb]{0.7,0,0}{+0.8}  \\

\midrule
Perplexity Guidance - 50\%	&16.4 &	20.6 &	39.1 &	37.1 	&28.3 	&16.7 &	16.4 	&8.2 &	26.0 &	16.8 &	29.3 	&25.1 &	25.2 &	19.1 &	24.7 \\
CaR -  50\%	&12.1 &	18.1 &	40.4 &	30.4 &	25.3 &	17.3 	&15.1 	&9.0 &	26.3 	&16.9  &	28.3& 	23.6 &	25.1 &	18.9 &	24.0   \\
Cherry Selection - 50\%	&15.5 &	19.5 	&38.9 &	37.3 	&27.8& 	15.4 &	16.3 &	8.8 &	26.1 &	16.7  &	\textbf{30.6} 	&24.8 &	25.3 	&18.9 &	24.9  \\
\rowcolor{blue!5} \textbf{GATEAU-LLaMA - 50\%}	& \textbf{18.5} 	& \textbf{22.5} 	& \textbf{43.9} 	& \textbf{39.1} 	& \textbf{31.0} 	& \textbf{17.9} 	& \textbf{16.7} 	& \textbf{9.6} 	& \textbf{28.0} 	& \textbf{18.1} 	& 30.1 	& \textbf{25.3} 	& \textbf{26.6} 	& \textbf{19.4} 	& \textbf{25.3}   \\
\rowcolor{blue!5}  $\Delta$ compared to Full - 100\%  &  \textcolor[rgb]{0.7,0,0}{+3.8} 	&  \textcolor[rgb]{0.7,0,0}{+2.4} 	&  \textcolor[rgb]{0.7,0,0}{+6.9} 	&  \textcolor[rgb]{0.7,0,0}{+2.1} 	&  \textcolor[rgb]{0.7,0,0}{+3.8} 	&  \textcolor[rgb]{0.7,0,0}{+2.5} 	&  \textcolor[rgb]{0.7,0,0}{+2.9} 	&  \textcolor[rgb]{0.7,0,0}{+1.0} 	&  \textcolor[rgb]{0.7,0,0}{+1.3} 	&  \textcolor[rgb]{0.7,0,0}{+1.9} 	&  \textcolor[rgb]{0.7,0,0}{+0.8} 	&  \textcolor[rgb]{0.7,0,0}{+0.8} 	&  \textcolor[rgb]{0.7,0,0}{+0.9} 	&  \textcolor[rgb]{0.7,0,0}{+0.8} 	&  \textcolor[rgb]{0.7,0,0}{+0.8}  \\
\bottomrule
\end{tabular}
}
\caption{Automatic metrics evaluation results (\%) on LongBench in two different settings. 
We use the ID to represent the dataset in LongBench, e.g., 1-1 is the ID of the NarrativeQA dataset. 
}
\label{tb:longbench-auto}
\end{table*}

\section{Evaluations}
\label{appendix:evaluations}

\subsection{LongBench-Chat}

\textbf{Evaluation Data. }
LongBench-Chat focuses on assessing LLMs’ instruction-following capability under the long context.
LongBench-Chat includes 50 long context real-world queries ranging from 10k to 100k in length.
It covers diverse aspects of instruction-following abilities such as reasoning, coding, summarization, and multilingual translation over long contexts.
It consists of 40 tasks in English and 10 in Chinese.
GPT-4 \citep{GPT-4} is employed to give a score on a scale of 10 to the machine-generated responses based on the annotated ground-truths.
\citet{bai2024longalignrecipelongcontext} finds that with their proposed few-shot evaluation prompting, GPT-4’s correlation with human annotations not only aligns but also surpasses the level of agreement among human annotators.

\noindent
\textbf{Evaluation Prompts.}
\
LongBench-Chat employs GPT-4 to score the model’s response in 1-10 based on a given human-annotated reference answer and few-shot scoring examples for each question.
We use the same prompt as LongBench-Chat to get GPT-4’s evaluation shown in Figure~\ref{fig:longbench_chat_evaluation}.

\subsection{LongBench}
\label{appendix_longbench}
\textbf{Evaluation Data. }
\
LongBench is the first bilingual, multitask benchmark tailored for long context understanding.
LongBench includes different languages (Chinese and English) to provide a more comprehensive evaluation of the large models’ bilingual capabilities in long-context understanding.
Detailed statistics of the used dataset in LongBench can be found in Table \ref{table:longbench-stat}.

\noindent
\textbf{Evaluation Prompts. }
We conduct GPT-4 evaluation for LongBench as \citet{bai2024longalignrecipelongcontext}.
As aligned models generally produce longer responses, rather than relying solely on the original automatic metrics (e.g., ROUGE) to evaluate the models’ replies, we employ GPT-4 to assess the model outputs based on their alignment with the ground-truth answers on LongBench.
For the first two QA tasks, the prompt for the GPT-4 evaluator is the same as \citet{bai2024longalignrecipelongcontext}, shown in Figure~\ref{fig:longbench_evaluation_qa}.
The prompt for GPT-4 evaluation on summarization tasks is the same as \citet{bai2024longalignrecipelongcontext}, shown in Figure~\ref{fig:longbench_evaluation_summarization}.

\noindent
\textbf{Automatic Metrics Evaluation Results.} \ We show the detailed 
automatic metric evaluation results on LongBench in Table \ref{tb:longbench-auto}.
Meanwhile, we can see that using 30\% of the whole long instruction-following dataset (\textit{GATEAU-LLaMA-30\%}) can achieve the best performance of LongBench in two different settings.
This is because of its ability to maintain an optimal balance between the volume and quality of the long instruction-following samples it utilizes, leading to the most desirable results.

\begin{figure*}
    \centering
    \includegraphics[width=0.6\textheight]{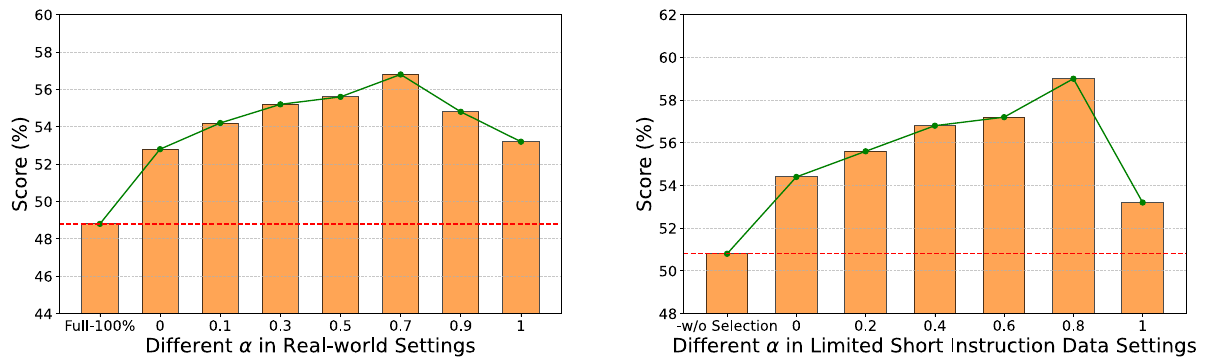}
    \caption{
    Results (\%) on LongBench-Chat with different hyperparameter $\alpha$ in Eq. (\ref{final_score}).
    }
    \label{fig_alpha}
\end{figure*}

\begin{figure}
    \centering
    \includegraphics[width=0.3\textheight]{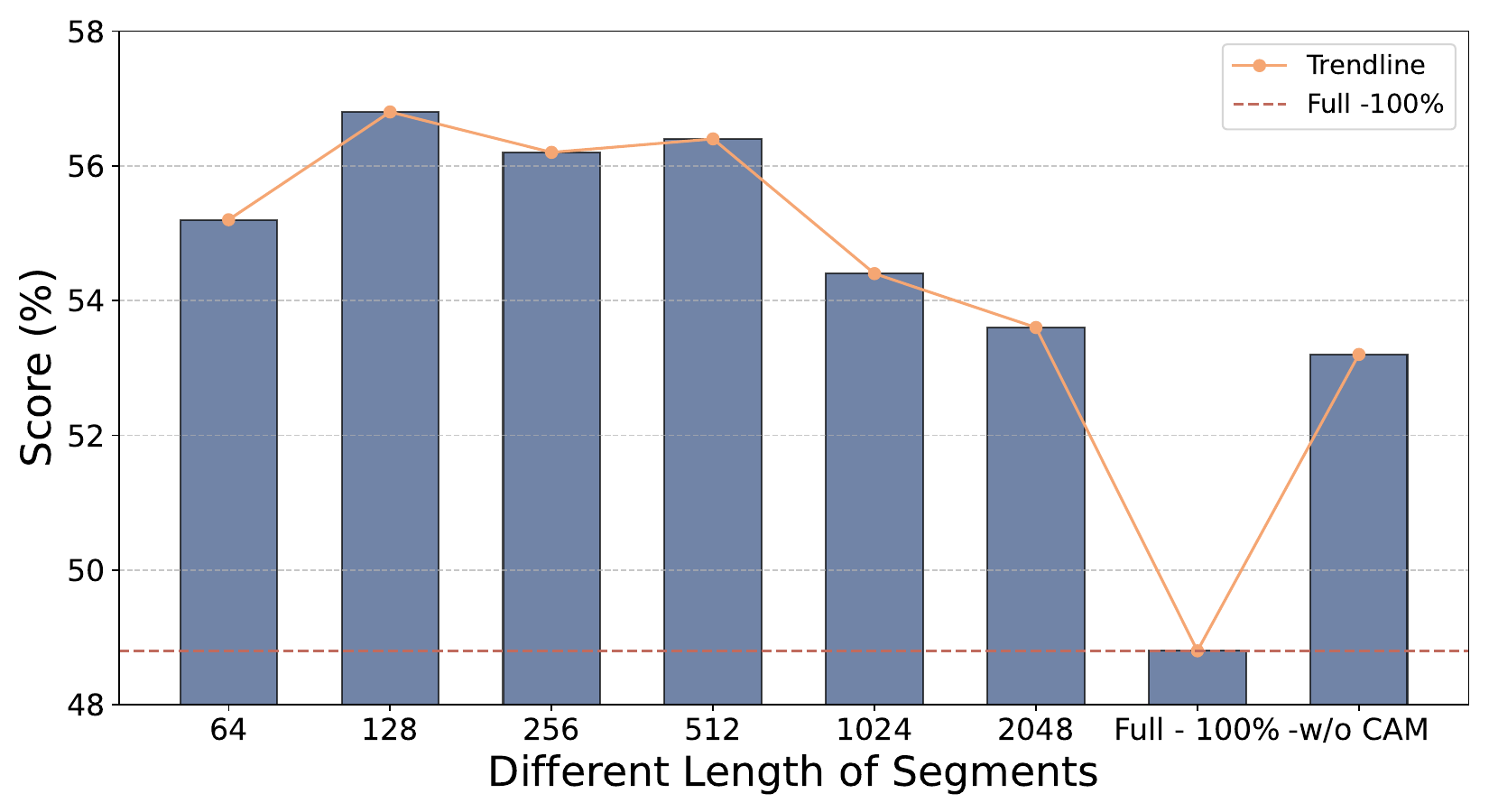}
    \caption{
    Results (\%) on LongBench-Chat with different hyperparameter $L$ in CAM.
    }
    \label{fig_length_appendix}
\end{figure}

\subsection{MT-Bench}
\textbf{Evaluation Data.}
\
MT-Bench is a comprehensive benchmark comprising 80 multi-turn questions. It is designed to assess the ability to engage in multi-turn conversations and follow instructions. The benchmark covers common use cases and emphasizes challenging questions to effectively differentiate among models.
It is meticulously designed to distinguish chatbots based on their fundamental capabilities, which include writing, roleplay, extraction, reasoning, mathematics, coding, knowledge in STEM fields, and knowledge in the humanities and social sciences.
MT-Bench prompts large language models, such as GPT-4, to serve as judges and evaluate the quality of the models' responses.
\citet{zheng2023judgingllmasajudgemtbenchchatbot} conducted a series of experiments and found that LLM judges like GPT-4 can align impressively well with both controlled and crowd-sourced human preferences, achieving over 80\% agreement. 
For each turn, GPT-4 assigns a score on a scale of 1 to 10. 
We then report the average score across all turns.

\noindent
\textbf{More Detailed Results.}
We show the detailed results of MT-Bench in Table \ref{tb:detailed_mtbench}.
\begin{table*}
\centering  
\resizebox{\textwidth}{!}{
\begin{tabular}{lccccccccccccccc}
\toprule
\textbf{Model} & \textbf{First-turn} & \textbf{Second-turn}  & \textbf{Writing} & \textbf{Roleplay} & \textbf{Reasoning} & \textbf{Math} & \textbf{Coding} & \textbf{Extraction} & \textbf{STEM} & \textbf{Humanities}
\\
\midrule
\multicolumn{16}{c}{\cellcolor{mypink}\textbf{
Real-world Settings}} \\

w/o SFT & 43.5&	25.6	&44.5&	44.0&	35.0	&16.5	&18.0	&28.0&	42.0&	48.8 \\
w/o Long SFT &60.0	&47.4&	73.8	&72.0&	44.0	&22.0	&25.5	&42.5	&63.0	&86.5
									\\
Full - 100\% &60.0	&48.7	&78.5&	70.3	&45.5	&19.0	&29.0	&42.0	&67.5	&83.0
									\\
         \midrule
Perplexity Guidance - 10\% &63.1	&48.9	&68.7&	67.0&	43.5	&26.5&	33.2&	50.5&	69.8	&88.5
									\\
CaR -  10\% &59.8&	50.0	&76.5	&75.3	&44.5	&24.5	&24.8	&43.5	&64.2&	84.9
									\\
Cherry Selection - 10\% &63.0	&50.5&	74.5	&73.8	&42.3&	25.0	&32.5	&48.3	&70.3&	87.5
									\\
\rowcolor{blue!5} \textbf{GATEAU-LLaMA - 10\%}  	& \textbf{63.1} 	& \textbf{54.1} 	&73.8 	&\textbf{79.2} 	&43.8 	&\textbf{26.5} 	&27.8 	&46.0 &	\textbf{77.0} 	&\textbf{94.8} 
										\\
Perplexity Guidance - 30\% &62.1	&47.8&	69.0&	63.7	&46.0	&28.0&	28.4	&49.0	&72.5	&82.2
									\\
CaR -  30\% &60.0	&48.6	&79.3	&77.0	&38.5&	21.0	&19.8	&44.0&	71.9&	83.0
									\\
Cherry Selection - 30\% &61.6	&47.0	&68.2&	71.5	&39.8	&22.0&	26.3	&50.8&	69.3	&88.4
									\\
\rowcolor{blue!5} \textbf{GATEAU-LLaMA - 30\%}& \textbf{64.1} 	&\textbf{50.4} 	&78.0 &	73.5 &	42.0 	&24.5 &	\textbf{29.5} &	46.8 &	\textbf{73.8} &	\textbf{92.1} 
									\\
Perplexity Guidance - 50\%& 62.3	&49.6&	79.0	&71.0	&47.3&	24.5	&28.0&	42.0&	69.5	&86.3 \\
CaR -  50\% &61.6	&47.9&	74.0	&77.3&	39.0&	21.5&	24.5&	42.0&	67.8&	91.8
									\\
Cherry Selection - 50\% &62.9	&49.6&	77.8	&76.2	&48.3&	22.5	&30.5&	35.8	&68.2	&91.5
									\\
\rowcolor{blue!5} \textbf{GATEAU-LLaMA - 50\%} &\textbf{64.1} &	\textbf{50.4} &	\textbf{78.0} &	73.5& 	42.0 &	\textbf{24.5} &	29.5 	&\textbf{46.8}& 	\textbf{73.8} 	& \textbf{92.1} 
									\\

\midrule
\multicolumn{16}{c}{\cellcolor{mypink}\textbf{
Limited Short Instruction Data Settings}} \\

w/o SFT &43.5 &	25.6 	&44.5& 	44.0 &	35.0 	&16.5 &	18.0 &	28.0 	&42.0 &	48.8
									\\
w/o Long SFT &56.4 &	44.5& 	66.3 &	65.8 &	46.5 	&21.0 	&23.5 	&38.3 &	63.5 	&79.1 
									\\
Full - 100\% &54.5 &	40.9 &	65.8 	&56.0 	&35.5 &	21.0 	&23.5 	&34.0 	&67.5 &	78.3 
									\\
\midrule
Perplexity Guidance - 10\% &61.9 &	39.5 	&73.8 &	61.8 &	39.3 &	27.5 &	29.1 &	47.1 &	58.5 &	72.3 
									\\
CaR -  10\% &59.3 &	40.3 &	66.5 &	64.3 	&49.3 &	21.5 &	26.3 &	28.8 &	62.0 &	80.5 
									\\
Cherry Selection - 10\% &53.0 	&42.3 &	56.8 	&72.3& 	39.5 	&17.0 &	26.5 &	34.8 	&59.3 &	75.3 
									\\
\rowcolor{blue!5} \textbf{GATEAU-LLaMA - 10\%} & \textbf{62.2} 	&\textbf{44.6} &	69.9 &	67.5 	&39.8& 	24.0 &	27.5 &	\textbf{50.7} 	&\textbf{66.3} &	\textbf{83.0} 
									\\
Perplexity Guidance - 30\% &58.9 	&41.4 	&69.4& 	68.0 	&37.0 	&28.5 &	28.9 	&47.8 &	57.8 	&64.8 
									\\
CaR -  30\% &52.8 &	44.3 &	67.0 	&66.5 	&37.3 &	25.0 &	24.8 &	28.5 &	68.5 &	71.0 
									\\
Cherry Selection - 30\% & 54.8& 	36.6 	&67.5 &	57.5 &	34.0 &	19.5 &	20.4 &	35.5 &	63.5 &	69.7 
									\\
\rowcolor{blue!5} \textbf{GATEAU-LLaMA - 30\%}  	&\textbf{62.0} &	43.7 	&62.0 	&65.7 &	\textbf{45.4} &	27.5 &	\textbf{31.7} &	41.7 	&\textbf{71.7} &	\textbf{72.0} 
									
									\\
Perplexity Guidance - 50\% & 57.6 &	40.9 &	59.5 &	74.5 &	41.0 &	25.0 &	26.0 &	37.3 &	55.3 	&75.3 \\
CaR -  50\% &58.3 &	44.1 &	70.0 	&67.2& 	43.3& 	25.5 &	30.5 	&28.5 	&71.5& 	73.5 
									\\
Cherry Selection - 50\% &57.7 &	41.4 &	70.0 &	63.2 	&37.5 	&18.3 	&26.3 	&43.9 &	61.1 	&76.5 
									\\
\rowcolor{blue!5} \textbf{GATEAU-LLaMA - 50\%}& \textbf{64.2} 	&\textbf{44.1} &	61.5 &	67.0 &	\textbf{46.3} &	\textbf{28.0} &	\textbf{31.4} &	\textbf{47.0} 	&65.8 &	\textbf{84.3} 
									\\

\bottomrule
\end{tabular}
}
\caption{Detailed results (\%) of MT-Bench. 
}
\label{tb:detailed_mtbench}
\end{table*}

\subsection{GPT-4 Version}
\label{sec_gpt}
For all the evaluations using the GPT-4 (including LongBench-Chat, LongBench, MT-Bench, and Needle in the Haystack test), we used \texttt{GPT-4 API} in August 2024 to ensure that we keep the same as \citet{bai2024longalignrecipelongcontext}. 
According to the documents from OpenAI, \texttt{GPT-4 API} points to \texttt{GPT-4-0613 API}.


\section{General Characteristics of Selected Samples from GATEAU}
\label{char_section}
Utilizing GPT-4, we evaluate each sample on five aspects: the coherence of long input contexts, the necessity of long input contexts, helpfulness of response, the faithfulness of response, and the complexity of instruction.
Different from the previous GPT-4 evaluation detailed in Appendix \ref{sec_gpt}, we use \texttt{GPT-4-Turbo API} (now points to \texttt{GPT-4-Turbo-2024-04-09}) as our evaluator, as this version of API has larger context window to conduct the more correct evaluation for our long input contexts.
To ensure stable evaluation results, we use GPT-4 to score twice on 200 selected samples, and then average these scores to obtain the final results.
The prompt for GPT-4 evaluation on different characteristics can be found in Figure~\ref{fig:coherence}, Figure~\ref{fig:long_input_contexts_evaluation}, Figure~\ref{fig:faithfulness_of_response_evaluation}, Figure~\ref{fig:helpfulness_of_response_evaluation}, and Figure~\ref{fig:complexity_of_instruction_evaluation}.





\section{Parameter Study}
\label{appendix:parameter}
As shown in Figure \ref{fig_alpha}, we conduct experiments to explore the impact of the important hyperparameter $\alpha$ in Eq. (\ref{final_score}).
We report the results of \textit{GATEAU-LLaMA - 50\%} on LongBench-Chat in two settings.
Overall, although the choice of different $\alpha$ will have some impact on the LLM’s performance, the performance will always be improved over the baseline \textit{Full-100\%}, i.e., using the whole training dataset without data selection.
Meanwhile, we also find that using both the Homologous Model’s Guidance and Contextual Awareness Measurement will further improve the performance than only using one of them.
This is because the Homologous Model’s Guidance and Contextual Awareness Measurement attempts to measure the difficulty brought by the long-range dependencies from two different perspectives, i.e., separately measuring the difficulty of generating corresponding responses and understanding long input contexts due to the long-range dependencies.
Meanwhile, we further explore the impact of the length of the segment $L$ in CAM.
We report the results of \textit{GATEAU-LLaMA - 50\%} on LongBench-Chat in Real-world Settings.
As shown in Figure \ref{fig_length_appendix}, different segment lengths affect the model's performance; however, as long as a reasonable length value is chosen, the fluctuations in model performance are not significant. Meanwhile, the performance will always be improved over using the whole long SFT dataset (namely \textit{Full-100\%}) and only using the HMG method (namely \textit{-w/o CAM}), showing the effectiveness of our proposed CAM.

\section{OOD Case Study}
\label{appendix:case}
As part of our research on aligning LLMs on long context, we further come up with a practical case study shown in Figure~\ref{fig:casestudy}.
We use an out-of-distribution (OOD) query, which has not been encountered in the long context SFT data.
Specifically, we select the \textit{Biden-Trump debate transcript \footnote{https://edition.cnn.com/2024/06/27/politics/read-biden-trump-debate-rush-transcript/index.html}} from the 2024 election season as the OOD query, because this debate is organized subsequent to the collection of our used training datasets.
We show the results generated by \textit{GATEAU-LLaMA-30\%} and \textit{Full-100\%} in Real-world settings.
We highlight the sentences that can be easily misunderstood or contain factual errors (e.g., this debate is organized in 2024 instead of 2020).
We can find that our method achieves better faithfulness and fluency.

\section{Discussion}
\label{appendix:dis}

\subsection{Discussion about Execution Time and GPU Burdens}
\label{appendix:dis_time}
\noindent
\textbf{Execution Time.}
\
Based on the principle of making full use of GPU devices (e.g., using a multi-processing strategy), we list the execution time in Table \ref{appendix_tb_time}.
We can find that GATEAU introduces an acceptable offline time overhead compared to the supervised fine-tuning stage and improves the overall performance of long-context LLMs. 
Perplexity Guidance applies a single LLM to compute the score, thus, it achieves less execution time but worse performance in our experiments. 
Meanwhile, another strong baseline Cherry Selection introduces an additional training stage and computes the proposed Instruction-Following Difficulty (IFD) by applying the forward propagation twice on a single long SFT data, thus necessitating more execution time compared to our proposed HMG. 
Meanwhile, our CAM and HMG can process the data in parallel to further decrease the execution time, e.g., only 8 hours with 16xA800 80G GPUs. 
The experimental results of our proposed GATEAU demonstrate that the additional execution time is worthwhile.

\begin{table*}[t]
\centering  
\scriptsize
\resizebox{\textwidth}{!}{
\begin{tabular}{lccccccccccccccc}
\toprule
\multirow{2}{*}{\textbf{Model}} & \multicolumn{3}{c|}{\textbf{LongBench}} & \multicolumn{1}{c|}{\textbf{LongBench-Chat}} & \multicolumn{3}{c}{\textbf{MT-Bench}} \\
\cmidrule(lr){2-4} \cmidrule(lr){5-5} \cmidrule(lr){6-9} 
& \textbf{Single-Doc QA} & \textbf{Multi-Doc QA} & \textbf{Summarization} & \textbf{Avg} & \textbf{First-turn} & \textbf{Second-turn} & \textbf{Avg} 
\\
\midrule
\multicolumn{16}{c}{\cellcolor{mypink}\textbf{
Real-world Settings}} \\
\rowcolor{blue!5} \textbf{GATEAU-ChatGLM3 - 6B - 50\%} & \textbf{30.2}   & \textbf{20.0}   & \textbf{24.6}   & \textbf{60.2} & \textbf{63.2}	& \textbf{50.2}	& \textbf{56.7} \\
\ -w/o Data Selection (i.e., Full - 100\%) & 26.6  & 16.9   & 23.4   &55.9&59.2	&47.6&	53.4\\
\rowcolor{blue!5} \textbf{GATEAU-LLaMA3 - 8B - 50\%} & \textbf{42.1} & \textbf{30.2} & \textbf{26.1} & \textbf{65.3} & \textbf{72.8} & \textbf{58.4} & \textbf{65.6}  \\
\ -w/o Data Selection (i.e., Full - 100\%) & 35.2 & 24.3 & 25.6 & 54.4 & 67.2 & 54.2 & 60.7 \\
    
\midrule
\multicolumn{16}{c}{\cellcolor{mypink}\textbf{
Limited Short Instruction Data Settings}} \\

\rowcolor{blue!5} \textbf{GATEAU-ChatGLM3 - 6B - 50\%} & \textbf{24.4}  & \textbf{15.4}  & \textbf{22.4}  & \textbf{57.0} & \textbf{57.2} &	\textbf{49.2} &	\textbf{53.2}  \\
\ -w/o Data Selection (i.e., Full - 100\%) &20.2 & 13.2 & 21.2 & 50.4& 55.4 &	45.2 	&50.3 \\
\rowcolor{blue!5} \textbf{GATEAU-LLaMA3 - 8B - 50\%} & \textbf{34.2} & \textbf{25.3} & \textbf{25.2} & \textbf{63.2} & \textbf{68.4} & \textbf{55.2} & \textbf{61.8} \\
\ -w/o Data Selection (i.e., Full - 100\%) & 30.2 & 23.2 & 24.7 & 55.6 & 62.3 & 50.2 & 56.3 \\

\bottomrule
\end{tabular}
}
\caption{Results (\%) of GATEAU-ChatGLM3 and GATEAU-LLaMA3 series. We show automatic metrics evaluation results on LongBench. 
}
\label{tb:appendix_glm}
\end{table*}

\begin{table*}[t]
\centering  
\scriptsize
\resizebox{\textwidth}{!}{
\begin{tabular}{lccccccccccccccc}
\toprule
\multirow{2}{*}{\textbf{Model}} & \multicolumn{3}{c|}{\textbf{LongBench}} & \multicolumn{1}{c|}{\textbf{LongBench-Chat}} & \multicolumn{3}{c}{\textbf{MT-Bench}} \\
\cmidrule(lr){2-4} \cmidrule(lr){5-5} \cmidrule(lr){6-9} 
& \textbf{Single-Doc QA} & \textbf{Multi-Doc QA} & \textbf{Summarization} & \textbf{Avg} & \textbf{First-turn} & \textbf{Second-turn} & \textbf{Avg} 
\\
\midrule
\multicolumn{16}{c}{\cellcolor{mypink}\textbf{
Real-world Settings}} \\
\rowcolor{blue!5} \textbf{GATEAU-LLaMA - 7B - 50\%} & \textbf{39.1}   & \textbf{27.5}   & \textbf{27.8}   & \textbf{50.2} & \textbf{55.7}	& \textbf{45.3}	& \textbf{50.5} \\
\ -w/o Data Selection (i.e., Full - 100\%) & 37.5  & 24.5   & 26.9   & 45.6 & 52.5	& 42.1 &	47.3 \\
		
\midrule
\multicolumn{16}{c}{\cellcolor{mypink}\textbf{
Limited Short Instruction Data Settings}} \\

\rowcolor{blue!5} \textbf{GATEAU-LLaMA - 7B - 50\%} & \textbf{32.5}   & \textbf{19.2}   & \textbf{26.4}   & \textbf{54.2} & \textbf{50.0}	& \textbf{42.8}	& \textbf{46.4} \\
\ -w/o Data Selection (i.e., Full - 100\%) & 28.4  & 17.0   & 25.5   & 48.2 & 47.5	& 41.4 &	44.5 \\

\bottomrule
\end{tabular}
}
\caption{Experiments to explore whether GATEAU can fit in other long SFT datasets.
We use LongAlpaca as the long SFT dataset.
We show automatic metrics evaluation results on LongBench.
}
\label{appendix_tb_datasets} 
\end{table*}

\begin{table}
\scriptsize	
\centering
\begin{tabular}{lc}
\\
\toprule
\textbf{Methods} & \textbf{Execution Time} \\
\midrule
\multicolumn{2}{c}{\cellcolor{mypink} \textbf{Real-world Settings}} \\
Training on the full dataset & 176 GPU hours \\
Selecting long SFT data via HMG	& 64 GPU hours \\
Selecting long SFT data via CAM & 48 GPU hours \\
Selecting long SFT data via Cherry Selection & 80 GPU hours \\
Selecting long SFT data via Perplexity Guidance & 32 GPU hours \\
\bottomrule
\end{tabular}
\caption{Execution time.}
\label{appendix_tb_time} 
\end{table}

\noindent
\textbf{GPU Burdens.}
GATEAU is designed to score long SFT data and then select the influential samples used for alignment.
Thus, our method does not introduce the additional memory burden during the SFT stage and inference stage.
For HMG, we compute perplexity scores generated from two models for a given SFT data in parallel and use the computed perplexity scores (cached in JSON files) to get the HMP score as shown in Eq. (\ref{eq_norm}). 
Thus, HMG does not introduce additional GPU memory burden, only introducing acceptable additional execution time as shown in Table \ref{appendix_tb_time}. 
The GPU memory requirements of CAM rise from the calculation of the attention scores for lengthy inputs, as well as the perplexity score computation.
This process is equivalent to performing two forward passes over the dataset without updating gradients, thus it does not add an extra GPU memory burden.

\subsection{Discussion about Whether GATEAU Can Fit in Other LLMs}
\label{appendix:dis_other}
We explore whether GATEAU can fit in other LLMs in Table \ref{tb:appendix_glm}.
We further apply GATEAU on ChatGLM3-6B-base-8k \citep{zeng2023glmb} and ChatGLM3-6B-base-64k \citep{bai2024longalignrecipelongcontext}, then fine-tune ChatGLM3-6B-base-64k using the selected samples.
We also conduct the experiments on LLaMA3-8B-base-8k and LLaMA3-8B-base-64k, then fine-tune LLaMA3-8B-base-64k using the selected samples.
We can find consistent improvements on three benchmarks compared to using the full long SFT dataset.
This indicates that GATEAU effectively fits in other LLMs.

\subsection{Discussion about Whether GATEAU Can Fit in Other Long SFT Datasets}
\label{appendix:dis_other_datasets}
Meanwhile, we explore whether GATEAU can fit in other long SFT datasets.
Specifically, we implement our proposed GATEAU on the long SFT dataset LongAlpaca \citep{chen2024longlora}, which contains 9,000 long SFT samples.
As shown in Table \ref{appendix_tb_datasets}, we can find that our method GATEAU achieves consistent improvements on three benchmarks, including long-context understanding benchmark and two instruction-following benchmarks, showing the GATEAU can generalize across different long SFT datasets.

\subsection{Discussion about the Diversity of Selected Samples}
\label{appendix:dis_diversity}
In this section, we further explore the diversity of selected samples.
We employ the cluster model as CaR~\citep{ge2024clustering} to cluster all candidate instruction pairs into $k$ clusters. 
Specifically, we employ the k-Means algorithm and a sentence transformer model, which is used to map sentences to a 384-dimensional dense vector space. 
Subsequently, semantic features are PCA-reduced to retain 95\% of dimensions. 
Finally, by setting the number of clusters as $k =  \sqrt{n/2}$ for $n$ long SFT samples, all 10k long SFT samples are clustered into 70 clusters. 
Finally, all samples are sorted based on their scores according to Eq. (\ref{final_score}), and the top $n_1$ samples are selected. 
Within each cluster, samples are sorted by score from GATEAU, and the top $n_2$ pairs are chosen. We set $n_2$ to 1, which is the same as \citet{ge2024clustering}. Finally, we can get $n_1 + k*n_2$ (i.e., $4300 + 70 * 1$) samples and use these selected data to train the model, namely \textit{-w Diversity-preserved Selection}. 
We report the results of GATEAU-LLaMA - 50\% on LongBench-Chat and MT-Bench.
Shown in Table \ref{appendix_tb_diversity}, we find that using \textit{Diversity-preserved Selection} does not consistently improve the performance, showing our proposed GATEAU has implicitly ensured the diversity of selected long SFT data.
This is because HMG and CAM separately measure the difficulty of generating corresponding responses and understanding long input contexts due to the long-range dependencies, thus the final score derived from two different perspectives inherently ensures the diversity of selected long SFT data.
Meanwhile, as shown in Table \ref{tb:detailed_mtbench}, GATEAU achieves better overall performance and more balanced performance in 8 different tasks, showing its effectiveness and diversity of selected samples.

\begin{table*}[t]
\centering  
\resizebox{\textwidth}{!}{
\begin{tabular}{lccccccccccccccc}
\toprule
\multirow{2}{*}{\textbf{Model}} & \multicolumn{3}{c|}{\textbf{LongBench}} & \multicolumn{1}{c|}{\textbf{LongBench-Chat}} & \multicolumn{3}{c}{\textbf{MT-Bench}} \\
\cmidrule(lr){2-4} \cmidrule(lr){5-5} \cmidrule(lr){6-9} 
& \textbf{Single-Doc QA} & \textbf{Multi-Doc QA} & \textbf{Summarization} & \textbf{Avg} & \textbf{First-turn} & \textbf{Second-turn} & \textbf{Avg} 
\\
\midrule
\multicolumn{16}{c}{\cellcolor{mypink}\textbf{
Real-world Settings}} \\
\rowcolor{blue!5} \textbf{GATEAU-LLaMA - 7B - 50\%} &\textbf{38.9}&\textbf{25.8}&25.5&\textbf{56.8} &\textbf{64.1}&\textbf{50.4}&\textbf{57.3}   \\
-w/o Extended Context Windows & 38.1 & 25.4 & \textbf{25.6} & 55.8
& 63.7	& 50.6 & 57.1 \\
-w/o Norm in Eq. (\ref{eq_norm}) & 37.5  & 24.1  & 25.3  & 56.2 &64.1 	& 50.4 & 57.3  \\
Homologous Model’s Guidance &38.4 &24.3&25.1&53.2 &61.7	&51.5 &56.6   \\
Perplexity Guidance & 37.9  & 23.4 & 25.4 & 49.8 &62.3&49.6&55.9\\
Non-Homologous Model’s Guidance  & 37.2 & 23.2 & 24.8 & 48.2 & 59.2 & 49.3 & 54.3\\
\midrule

\multicolumn{16}{c}{\cellcolor{mypink}\textbf{
Limited Short Instruction Data Settings}} \\
\rowcolor{blue!5} \textbf{GATEAU-LLaMA - 7B - 50\%} & \textbf{31.0}  & 18.1  & \textbf{25.3}  & \textbf{59.0} & \textbf{64.2} &	\textbf{44.1} &	\textbf{54.2}  \\
-w/o Extended Context Windows &29.2& \textbf{18.8}& 25.2& 57.6&60.2 &	44.0 & 52.1 \\
-w/o Norm in Eq. (\ref{eq_norm}) & 29.7 & 18.7 & 24.9&  55.2&  62.0 	& 40.1 & 51.1 \\
Homologous Model’s Guidance & 28.5 & 17.5 & 24.7 &53.2&61.3 &	42.4 &51.8  \\
Perplexity Guidance & 28.3& 16.8 &24.7 &51.0&57.6 &	40.9 &49.2\\
Non-Homologous Model’s Guidance & 28.7 & 16.8 & 24.8 & 50.2 & 60.1 & 40.3 & 50.2 \\

\bottomrule
\end{tabular}
}
\caption{Discussion about Homologous Model’s Guidance.}
\label{tb:exp_hmg}
\end{table*}

\begin{table}[t]
\scriptsize	
\centering
\begin{tabular}{lcc}
\\
\toprule
\textbf{Models} & \textbf{LongBench-Chat}  &\textbf{MT-Bench} \\
\midrule
\multicolumn{3}{c}{\cellcolor{mypink} \textbf{Real-world Settings}} \\
GATEAU-LLaMA - 7B - 50\% & \textbf{56.8} & 57.3 \\
-w/ Diversity-preserved Selection & 	56.2 & \textbf{57.8} \\
\midrule
\multicolumn{3}{c}{\cellcolor{mypink} \textbf{Limited Short Instruction Data Settings}}\\
GATEAU-LLaMA - 7B - 50\% & 59.0 & \textbf{54.2} \\
-w/ Diversity-preserved Selection & \textbf{59.2} & 53.4 \\
\bottomrule
\end{tabular}
\caption{Experiments to explore the diversity of selected samples by GATEAU.}
\label{appendix_tb_diversity} 
\end{table}

\subsection{Discussion about Homologous Model’s Guidance}
We further explore some key questions in the Homologous Model’s Guidance.
\label{appendix:dis_hmg}

\noindent
\textbf{Why Do We Need Homologous Models?}
\
Homologous Model’s Guidance (HMG) aims to assess the degree of long-range dependencies required for the corresponding response generation by comparing the perplexity scores of the response between two homologous models with different context windows.
The idea behind HMG is that the primary difference between homologous models with varying context windows lies in their different capabilities for modeling long-range dependencies instead of other capabilities. 
Thus, the disparity in the perplexity scores can be interpreted as reflecting the difference in the long-range dependencies modeling capabilities required to generate the given response. 
To evaluate the effectiveness of our idea, we replace \textit{LLaMA-2-7B-base-4k} with \textit{Qwen-2-7b-base-8k} \citep{yang2024qwen2technicalreport} as model $\theta_{A}$ in Eq. (\ref{eq_norm}), namely \textit{Non-Homologous Model’s Guidance}.
As shown in Table \ref{tb:exp_hmg}, we find \textit{Non-Homologous Model’s Guidance} achieves worse performance than \textit{Homologous Model’s Guidance} in two designed settings.
It shows that HMG can exclusively measure the richness of long-range dependency relations in long SFT samples.
As non-homologous models have different pre-training phases and model architectures, the modified Eq. (\ref{eq_norm}) can not effectively measure the degree of long-range dependencies required for response generation and introduce the influence brought by other different capabilities of non-homologous models, resulting in worse performance.

\noindent
\textbf{Why Do We Apply Normalization in Eq. (\ref{eq_norm}) ?}
\
We apply softmax normalization to each score in Eq. (\ref{eq_norm}) to determine its respective ranking among the datasets for two perplexity scores.
This is because our early experiments observed that applying softmax normalization can further improve the performance shown in Table \ref{tb:exp_hmg}.
This may due to the fact that some extremely noisy samples tend to have large perplexity scores, which in turn lead to unstable HMP scores if we do not apply normalization in Eq. (\ref{eq_norm}).
Training LLMs on these noisy samples further leads to poor results.

\noindent
\textbf{What Will Happen if We Do not Extend the Context Windows of LLaMA-2-4k?}
\
Our early experiments also explore what will happen if we do not extend the context windows of model $\theta_{A}$ in Eq. (\ref{eq_norm}).
As shown in Table \ref{tb:exp_hmg}, we are surprised to find that \textit{-w/o Extended Context Windows} also achieves competitive results in three benchmarks compared to \textit{GATEAU-LLaMA}.
Even the perplexity score ${\rm PPL}_{\theta_A}(y|c,x)$ from the model $\theta_{A}$ can be very large, e.g., the value of ${\rm PPL}_{\theta_A}(y|c,x)$ can be larger than 1000, the value after softmax normalization is still useful and applicable in the Homologous Models’ Guidance.
This interesting finding can be used to reduce the complexity of applying Homologous Models’ Guidance and achieve competitive performance.

\begin{table}
\small
\centering
\begin{tabular}{lc}
\\
\toprule
\textbf{Models} & \textbf{Perplexity} \\
\midrule
LLaMA-2-7B-base-4k & 3.72 \\
LLaMA-2-7B-base-64k & 2.61 \\
\bottomrule
\end{tabular}
\caption{Perplexity from base models on LongAlign.}
\label{appendix_tb_ppl} 
\end{table}

\noindent
\textbf{Is the Perplexity Score from the Base Model Really so High that It Can Not Accurately Measure the Difficulty?}
\
As the base model performs well on conditional generation tasks like continuation, it should be able to generate accurate perplexity scores on the response of instruction-following data, even though the model might not be able to produce high-quality responses correctly, because these two capabilities are not the same. 
We explore whether our long-context LLM would produce incorrect perplexity values in Table \ref{appendix_tb_ppl}. 
We calculate the average perplexity value generated by LLaMA-2-7B-base-64k for the entire long SFT dataset LongAlign during the whole HMG process, which is 2.61. 
We further calculate the average perplexity value generated by LLaMA-2-7B-base-4k for the entire long SFT dataset LongAlign during the whole HMG process, which is 3.72.
This is because we expand the base frequency of the RoPE position encoding by 200 times (from 10,000 to 2,000,000) to extend the context windows and avoid the model conducting extreme perplexity score (e.g., >1,000) in HMG.
Thus, there is no issue of the perplexity from the base model being too high to accurately measure the difficulty.

\begin{table}[t]
\scriptsize	
\centering
\begin{tabular}{lcc}
\\
\toprule
\textbf{Models} & \textbf{LongBench-Chat}  &\textbf{MT-Bench} \\
\midrule
\multicolumn{3}{c}{\cellcolor{mypink} \textbf{Real-world Settings}} \\
GATEAU-LLaMA - 7B - 50\% & \textbf{56.8} & 57.3 \\
w/ ICL Alignment & 56.2 & \textbf{57.9} \\
\midrule
\multicolumn{3}{c}{\cellcolor{mypink} \textbf{Limited Short Instruction Data Settings}}\\
GATEAU-LLaMA - 7B - 50\% & 59.0 & \textbf{54.2} \\
w/ ICL Alignment & \textbf{59.4} & 53.5 \\
\bottomrule
\end{tabular}
\caption{Experiments to explore whether alignment via in-context learning helps HMG.}
\label{appendix_tb_icl} 
\end{table}

\noindent
\textbf{Can the Perplexity Score Generated from the Base Model be Used as Guidance to Select Influential Samples?}
\
The perplexity of the responses computed with the base model is an intuitive metric, as it measures the difficulty of the data sample during the generation. 
As shown in Table \ref{tb:longchat_mt}, Table \ref{tb:longbench-gpt4}, Table \ref{tb:longbench-auto}, and Table \ref{fig:longbench_chat_evaluation},  we find simply using high perplexity (namely Perplexity Guidance in our paper) can also improve the performance compared with using the whole long SFT dataset, indicating that the effectiveness of the perplexity score from the base model in selecting long SFT samples. 
Cherry Selection \citep{li-etal-2024-quantity} also finds using the Instruction-Following Difficulty (a variant of perplexity score) computed with the base model also works in selecting SFT samples. 
According to these experiments, we believe that the perplexity generated from a base model can be used as positive guidance to select SFT samples. 
Therefore, the use of the perplexity score generated from the base model in our method makes sense when selecting long SFT data. 
Meanwhile, our method HMG is designed to minimize the influence of other factors (e.g., the limited instruction-following ability of a base model) and model the difficulty in modeling the long-range dependencies to construct the more effective guidance of long SFT data selection, and further improve overall performance. 
For CAM, utilizing perplexity scores to compute importance scores is also reasonable, and the experiments show improvement even when only using CAM.

\begin{table*}[t]
\centering  
\resizebox{\textwidth}{!}{
\begin{tabular}{lccccccccccccccc}
\toprule
\multirow{2}{*}{\textbf{Model}} & \multicolumn{3}{c|}{\textbf{LongBench}} & \multicolumn{1}{c|}{\textbf{LongBench-Chat}} & \multicolumn{3}{c}{\textbf{MT-Bench}} \\
\cmidrule(lr){2-4} \cmidrule(lr){5-5} \cmidrule(lr){6-9} 
& \textbf{Single-Doc QA} & \textbf{Multi-Doc QA} & \textbf{Summarization} & \textbf{Avg} & \textbf{First-turn} & \textbf{Second-turn} & \textbf{Avg} 
\\
\midrule
\multicolumn{16}{c}{\cellcolor{mypink}\textbf{
Real-world Settings}} \\
\rowcolor{blue!5} \textbf{GATEAU-LLaMA - 7B - 10\%} &\textbf{40.5}&\textbf{24.0}&\textbf{25.4}&\textbf{55.4} &\textbf{63.1}&\textbf{54.1}&\textbf{58.6}   \\
-w/o Mixed-Quality Data & 39.1 & 22.4 & 24.6 & 53.6
& 61.8	& 52.3 & 57.1 \\

\midrule

\multicolumn{16}{c}{\cellcolor{mypink}\textbf{
Limited Short Instruction Data Settings}} \\
\rowcolor{blue!5} \textbf{GATEAU-LLaMA - 7B - 10\%} & \textbf{29.9} & \textbf{18.4} & \textbf{24.6}  & \textbf{58.0} & \textbf{62.2} &	\textbf{44.6} &	\textbf{53.4}  \\
-w/o Mixed-Quality Data & 28.7 & 17.5 & 24.2 & 55.6
& 61.2	& 42.3 & 51.8 \\

\bottomrule
\end{tabular}
}
\caption{Discussion about training models on mixed-quality data. }
\label{tb:mix}
\end{table*}

\begin{table}
\small
\centering
\begin{tabular}{lc}
\\
\toprule
\textbf{Methods} & \textbf{Data Overlap Ratio with GATEAU} \\
\midrule
Cherry LLM & 12\% \\
CaR & 5\% \\
\bottomrule
\end{tabular}
\caption{Data overlap of top 10\% of long SFT data selected by baselines and our proposed GATEAU.}
\label{appendix_tb_overlap} 
\end{table}

We further conduct additional experiments to explore the effect of perplexity scores generated from the base model. 
In HMG, we use in-context learning technology to align the base model and use the perplexity score from the aligned model to select long SFT data. 
Specifically, we use the same 3 demonstration examples as URIAL \citep{lin2024the}. 
In this way, we can get models more aligned without updating the parameters.
However, as shown in Table \ref{appendix_tb_icl}, using the aligned model via in-context learning does not consistently improve the final performance. 
This indicates that using only base models in the HMG phase can also achieve good results. 
Therefore, HMG can effectively minimize the influence of other factors (e.g., the limited instruction-following ability of a base model) and model the difficulty in modeling the long-range dependencies.
Meanwhile, from the real-world implementation viewpoint, directly using the base model is more efficient and at the same time effective as well.

\subsection{Discussion about Orthogonality with Baselines}
\label{appendix:dis:orth}
To provide additional evidence of the unique benefits of our approach, we conduct additional experiments to analyze the orthogonality between GATEAU and various baselines.
Specifically, we calculate the overlap of the top 10\% of long SFT data selected by other baselines and our method in Limited Short Instruction Data Settings. 
As shown in the Table \ref{appendix_tb_overlap}, we can find a significant difference between the samples selected by the baselines focused on short SFT data selection (i.e., Cherry LLM and CaR) and those selected by our proposed GATEAU. 
This is because GATEAU is designed to identify the influential long samples enriched with long-range dependency relations to achieve better long context alignment instead of focusing on selecting short SFT data. 
Thus, GATEAU grasps important patterns that differ from the existing baselines. Furthermore, how to utilize such orthogonality to improve the final performance remains a promising research topic. 
We attempt to explore how to utilize it to further improve the final performance in our future work.

\subsection{Discussion about Training Models on Mixed-Quality Data}
\label{appendix:dis-mix}
We further attempt to explore whether there exists an optimal balance between low-scoring and high-scoring long SFT samples that enables the long-context LLM to perform even better than using the samples with high scores.
Specifically, we use long SFT samples from the top 5\% and bottom 5\% to form the training samples, namely Mixed-Quality Data.
As shown in Table \ref{tb:mix}, we find that this strategy does not improve the final performance of the LLMs
This indicates that our strategy of using top-ranked samples is effective.

\section{Case Study for Characteristics of Selected Samples}
\label{appendix:csc}
We conduct a case study to show the difference between samples with high or low scores generated by GATEAU.
In Figure \ref{fig:rank_case}, we show the sample with the highest score and the sample with the lowest score in Real-world Settings.
We highlight the low-quality sentences.
We can find that the sample with the highest score shows better faithfulness and fluency, showing the effectiveness of our method in selecting high-quality influential samples for long context alignment.

\newpage
\begin{figure*}[t]
    \centering
    \begin{tcolorbox}[title = {LongBench-Chat Evaluation Prompt}, size=title, colframe = white, colbacktitle = black!65!white]
    [Instructions] 
    You are asked to evaluate the quality of the AI assistant's answers to user questions as an impartial judge, and your evaluation should take into account factors including correctness (high priority), helpfulness, accuracy, and relevance. The scoring principles are as follows:

    1. Read the AI assistant's answer and compare the assistant's answer with the reference answer.

    2. Identify all errors in the AI Assistant's answers and consider how much they affect the answer to the question.

    3. Evaluate how helpful the AI assistant's answers are in directly answering the user's questions and providing the information the user needs. 

    4. Examine any additional information in the AI assistant's answer to ensure that it is correct and closely related to the question. If this information is incorrect or not relevant to the question, points should be deducted from the overall score.

    Please give an overall integer rating from 1 to 10 based on the above principles, strictly in the following format: "[[rating]]", e.g., "[[5]]". \\

    [Question] \{\}

    [Reference answer begins] \{\}
    [Reference answer ends]

    Below are several assistants' answers and their ratings:

    [Assistant's answer begins] \{\}
    [Assistant's answer ends]
    Rating: [[\{\}]]

    [Assistant's answer begins] \{\}
    [Assistant's answer ends]
    Rating: [[\{\}]]

    [Assistant's answer begins] \{\}
    [Assistant's answer ends]
    Rating: [[\{\}]] \\

    Please rate the following assistant answers based on the scoring principles and examples above:

    [Assistant's answer begins] \{\}
    [Assistant's answer ends]

    Rating:
    \end{tcolorbox}
    \caption{LongBench-Chat evaluation prompt.}
    \label{fig:longbench_chat_evaluation}
\end{figure*}
\clearpage

\begin{figure*} 
    \centering
    \begin{tcolorbox}[title = {LongBench Evaluation Prompt for QA tasks}, size=title, colframe = white, colbacktitle = black!65!white]
    \noindent
    You are asked to evaluate the quality of the AI assistant's answers to user questions as an impartial judge, and your evaluation should take into account factors including correctness (high priority), and comprehensiveness (whether the assistant's answer covers all points). 
    Read the AI assistant's answer and compare it against the reference answer, and give an overall integer rating of 1, 2, or 3 (1 = wrong or irrelevant, 2 = partially correct, 3 = correct and comprehensive) based on the above principles, strictly in the following format: "[[rating]]", e.g., "[[2]]". \\

    Question: \{\emph{Question}\}

    Reference answer: \{\emph{Groundtruth}\}

    Assistant's answer: \{\emph{Response}\}

    Rating:
    \end{tcolorbox}
    \caption{LongBench evaluation prompt for QA tasks.}
    \label{fig:longbench_evaluation_qa}
\end{figure*}

\begin{figure*} 
    \centering
    \begin{tcolorbox}[title = {LongBench Evaluation Prompt for Summarization Tasks}, size=title, colframe = white, colbacktitle = black!65!white]
    \noindent
    You are asked to evaluate the quality of the AI assistant's generated summary as an impartial judge, and your evaluation should take into account factors including correctness (high priority), comprehensiveness (whether the assistant's summary covers all points), and coherence.
    Read the AI assistant's summary and compare it against the reference summary, and give an overall integer rating on a scale of 1 to 5, where 1 is the lowest and 5 is the highest based on the evaluation criteria, strictly in the following format: "[[rating]]", e.g., "[[3]]". \\

    Reference summary: \{\emph{Groundtruth}\}

    Assistant's summary: \{\emph{Response}\}

    Rating:
    \end{tcolorbox}
    \caption{LongBench evaluation prompt for summarization tasks.}
    \label{fig:longbench_evaluation_summarization}
\end{figure*}

\begin{figure*} 
    \centering
    \begin{tcolorbox}[title = {Evaluation Prompt for the Coherence of Long Input Contexts}, size=title, colframe = white,colbacktitle = black!65!white]
    \noindent
    You are asked to evaluate the Long Input Contexts as an impartial judge, and your evaluation should follow these scoring principles:
    
    1. Read the given Long Input Contexts carefully.  \\
    2. Evaluate the fluency and coherence of Long Input Contexts. \\
    3. Evaluate whether the Long Input Contexts are focused and relevant. \\
    
    Please give an overall integer rating from 1 to 5 based on the above principles, strictly in the following format:"[[rating]]", e.g. "[[5]]". \\ \\
    Please rate the following Long Input Contexts based on the scoring principles: \\
    
    [Long Input Contexts begins]
    
    \{\emph{Long Input Contexts}\}
    
    [Long Input Contexts ends] \\
    
    Rating:
    \end{tcolorbox}
    \caption{Evaluation prompt for the coherence of long input contexts.}
    \label{fig:coherence}
\end{figure*}

\begin{figure*} 
    \centering
    \begin{tcolorbox}[title = {Evaluation Prompt for the Necessity of Long Input Contexts}, size=title, colframe = white, colbacktitle = black!65!white]
    \noindent
    You are asked to evaluate the Long Input Contexts as an impartial judge, and your evaluation should follow these scoring principles:

    1. Read the given Instruction, Long Input Contexts and Assistant's answer carefully. \\
    2. Evaluate how difficult it is for the Assistant to follow the given Instruction without the given Long Input Contexts. \\ 
    3. Evaluate how necessary the given Long Input Contexts are to get the Assistant's answer. If the Long Input Contexts are meaningless or irrelevant, points should be deducted from the overall score. \\

    Please give an overall integer rating from 1 to 5 based on the above principles, strictly in the following format: "[[rating]]", e.g., "[[5]]". \\

    [Instruction begins]

    \{\emph{Instruction}\}

    [Instruction ends] \\

    [Long Input Contexts begins]

    \{\emph{Long Input Contexts}\}

    [Long Input Contexts ends] \\

    Please rate the following assistant answers based on the scoring principles: \\

    [Assistant's answer begins]

    \{\emph{Assistant's answer}\}

    [Assistant's answer ends] \\
      
    Rating:
    \end{tcolorbox}
    \caption{Evaluation prompt for necessity of long input contexts.}
    \label{fig:long_input_contexts_evaluation}
\end{figure*}

\begin{figure*} 
    \centering
    \begin{tcolorbox}[title = {Evaluation Prompt for the Faithfulness of Response}, size=title, colframe = white, colbacktitle = black!65!white]
    \noindent
    You are asked to evaluate the AI assistant's answers to user questions as an impartial judge, and your evaluation should follow these scoring principles:

    1. Read the given Instruction, Long Input Contexts, and Assistant's answer carefully. \\
    2. Identify all errors in the AI Assistant's answers and consider how much they affect the answer to the question. \\
    3. Evaluate how faithful the AI assistant's answers are to follow the Instruction, i.e., how correct and closely related to the Instruction.  \\
    4. Evaluate how faithful the AI assistant's answers are based on the Long Input Contexts, i.e., how correct and closely related to the Long Input Contexts.  \\

    Please give an overall integer rating from 1 to 5 based on the above principles, strictly in the following format: "[[rating]]", e.g., "[[5]]". \\

    [Instruction begins]

    \{\emph{Instruction}\}

    [Instruction ends] \\

    [Long Input Contexts begins]

    \{\emph{Long Input Contexts}\}

    [Long Input Contexts ends] \\

    Please rate the following assistant answers based on the scoring principles: \\

    [Assistant's answer begins]

    \{\emph{Assistant's answer}\}

    [Assistant's answer ends] \\

    Rating:
    \end{tcolorbox}
    \caption{Evaluation prompt for faithfulness of response.}
    \label{fig:faithfulness_of_response_evaluation}
\end{figure*}

\begin{figure*} 
    \centering
    \begin{tcolorbox}[title = {Evaluation Prompt for the Helpfulness of Response}, size=title, colframe = white, colbacktitle = black!65!white]
    \noindent   
    You are asked to evaluate the AI assistant's answers to user questions as an impartial judge, and your evaluation should follow these scoring principles:

    1. Read the given Instruction and Assistant's answer carefully. \\
    2. Identify all errors in the AI Assistant's answers and consider how much they affect the answer to the question. \\
    3. Evaluate how helpful the AI assistant's answers are in directly answering the user's questions and providing the information the user needs.  \\

    Please give an overall integer rating from 1 to 5 based on the above principles, strictly in the following format: "[[rating]]", e.g. "[[5]]". \\

    [Instruction begins]

    \{\emph{Instruction}\}

    [Instruction ends] \\

    Please rate the following assistant answers based on the scoring principles: \\

    [Assistant's answer begins]

    \{\emph{Assistant's answer}\}

    [Assistant's answer ends] \\

    Rating:
    \end{tcolorbox}
    \caption{Evaluation prompt for helpfulness of response.}
    \label{fig:helpfulness_of_response_evaluation}
\end{figure*}
\clearpage

\begin{figure*}
    \centering
    \begin{tcolorbox}[title = {Evaluation Prompt for the Complexity of Instruction}, size=title, colframe = white, colbacktitle = black!65!white]
    \noindent   
    You are asked to evaluate the Instruction as an impartial judge, and your evaluation should follow these scoring principles:

    1. Read the given Instruction carefully. \\
    2. Evaluate the scope of the Instruction, i.e., whether the Instruction encompasses information necessary for successful completion. \\
    3. Evaluate the depth of the Instruction, i.e., whether the Instruction provides thorough details and nuances. \\
    4. Evaluate whether the Instruction integrates multiple steps or concepts that require careful attention and understanding. \\ 
    5. If the Instruction is too easy to follow, points should be deducted from the overall score.

    Please give an overall integer rating from 1 to 5 based on the above principles, strictly in the following format: "[[rating]]", e.g., "[[5]]". \\

    Please rate the following Instruction based on the scoring principles and examples above: \\

    [Instruction begins]

    \{\emph{Instruction}\}

    [Instruction ends] \\

    Rating:
    \end{tcolorbox}
    \caption{Evaluation prompt for complexity of instruction.}
    \label{fig:complexity_of_instruction_evaluation}
\end{figure*}
\clearpage

\begin{figure*}
    \centering
    \begin{tcolorbox}[title = {The Principles of Human Evaluation}, size=title, colframe = white, colbacktitle = black!65!white]
    \noindent   
    You are asked to evaluate the responses generated by different models, and your evaluation should follow these principles:

    1. \textbf{Correctness}: Focus primarily on whether the response covers the key points in the reference answer. For reference answers containing multiple key points, look for how many of these the response accurately addresses and score accordingly. \\
    2. \textbf{Helpfulness}: Evaluate how helpful the AI assistant’s answers are in directly answering the user’s questions and providing the information the user needs. \\
    3. \textbf{Faithfulness}: Examine any additional information in the AI assistant’s answer to ensure that it is correct and closely related to the question. 
    If this information is incorrect or not relevant to the question, points should be deducted from the overall score. \\
    4. \textbf{Coherence}: Evaluate how fluent and coherent the AI assistant’s answers are. Also, consider deducting points for overly verbose responses or those that are excessively generalized. \\

    Finally, please make a decision among the 3 opinions, including Win, Tie, and Loss.

    \end{tcolorbox}
    \caption{The principles of human evaluation.}
    \label{fig:human_evaluation_principles}
\end{figure*}
\clearpage

\begin{figure*} 
    \small
    \centering
    \begin{tcolorbox}[title = {OOD Case Study}, size=title, colframe = white, colbacktitle = black!65!white]
    \noindent
    \

    \faUser~\textbf{User:} 

    [Debate transcript begins] \{\}

    [Debate transcript ends] 

    [Instruction begins]

    As a professional summarizer, create a concise and comprehensive summary of the provided debate, while adhering to these guidelines:

    1. Craft a summary that is detailed, in-depth, and complex, while maintaining clarity and conciseness.

    2. Use proper grammar, punctuation, and spelling. 

    3. Write in a formal, objective, and impartial tone. 

    4. Provide a clear and accurate summary of the debate, without bias or personal opinions. 

    [Instruction ends] 

    Summary: \\ \\

    \faLaughBeam~\textbf{Ours:} 
    The CNN presidential debate between President Joe Biden and former President Donald Trump was held in Atlanta, Georgia. 
    The debate was moderated by CNN anchors Jake Tapper and Dana Bash. 
    The debate covered various topics including the economy, immigration, foreign policy, democracy, and more.  
    On the economy, President Biden defended his administration's actions, stating that he inherited a disastrous situation from the previous administration. 
    He highlighted his administration's job creation efforts and infrastructure spending. 
    Former President Trump criticized Biden's policies, claiming that Biden's tax hikes and regulations were hurting the economy.  
    On immigration, President Biden defended his administration's handling of the border crisis, stating that he had increased border patrol and asylum officers. 
    Former President Trump criticized Biden's immigration policies, claiming that Biden was allowing "terrorists" to enter the country.  
    On foreign policy, President Biden criticized former President Trump's handling of foreign affairs, particularly in regards to Russia and Ukraine. 
    He stated that Trump had encouraged Putin to invade Ukraine and that Trump had not done enough to stop the war. 
    Former President Trump defended his actions, stating that he had put pressure on Ukraine to investigate Biden's son and that he had not done enough to stop the war because he was focused on China.  
    On democracy, President Biden criticized former President Trump's handling of the January 6th attack on the Capitol, stating that Trump had encouraged the protesters to go to the Capitol. 
    Former President Trump defended his actions, stating that he had spoken peacefully and patriotically.  
    Overall, the debate was contentious and both candidates strongly defended their records and policies. \\ \\

    \faFrown~\textbf{Full - 100\%:} 
    The debate between President Joe Biden and former President Donald Trump was held in Atlanta, Georgia, \textcolor[rgb]{0.7,0,0}{on October 22, 2020}. 
    The debate was moderated by CNN anchors Jake Tapper and Dana Bash. The debate focused on various issues such as the economy, immigration, foreign policy, democracy, and more.  
    On the issue of the economy, President Biden argued that he inherited a disastrous economy from Trump, with high unemployment rates and no jobs. 
    He claimed that his policies have helped create jobs and increase wages for working-class Americans. 
    Trump, on the other hand, claimed that he had the best economy in history and that Biden's policies have caused inflation and driven up prices. On the issue of immigration, Trump claimed that Biden was allowing "millions of people to come in and destroy our country." He also claimed that Biden was responsible for the worst border crisis in history. Biden argued that Trump had left the border in chaos and that his policies had caused the crisis.  
    \textcolor[rgb]{0.7,0,0}{On foreign policy, Trump claimed that he had brought the United States to the brink of war with Iran and that he had made a deal with North Korea that was a "disaster."} Biden argued that Trump had put the United States in a dangerous position with his rhetoric and that he had made the world a more dangerous place.  
    On democracy, Trump claimed that the election was rigged and that he had won the election. Biden argued that Trump had incited a violent insurrection at the Capitol and that he had undermined democracy.  
    In conclusion, the debate was contentious and both candidates presented their respective arguments and policies. However, it remains to be seen how the voters will interpret the debate and how it will impact the election outcome. 

    \end{tcolorbox}
    \caption{OOD case study.}
    \label{fig:casestudy}
\end{figure*}

\begin{figure*} 
    \centering
    \begin{tcolorbox}[title = {Case Study for Characteristics of Selected Samples}, size=title, colframe = white, colbacktitle = black!65!white]
    \noindent
    \

    \faLaughBeam~\textbf{Sample with the highest score:}  \\
    
    [Long Input Contexts begins] 
    
    \{Long Input Contexts\}
    
    [Long Input Contexts ends] \\
    
    Instruction: Since what year has SAS been giving customers around the world THE POWER TO KNOW®? \\
    
    Response: Based on the information in the doc, SAS has been giving customers around the world THE POWER TO KNOW® since 1976. Specifically, the last paragraph states: "SAS is the leader in business analytics software and services, and the largest independent vendor in the business intelligence market. Through innovative solutions, SAS helps customers at more than 65,000 sites improve performance and deliver value by making better decisions faster. Since 1976 SAS has been giving customers around the world THE POWER TO KNOW®." Therefore, the answer is that SAS has been giving customers THE POWER TO KNOW® since 1976. \\ \\ \\

    \faFrown~\textbf{Sample with the lowest score:} \\
    
    [Long Input Contexts begins] 
    
    \{Long Input Contexts\}
    
    [Long Input Contexts ends] \\

    \begin{CJK}{UTF8}{gbsn}
    Instruction: Bookertainment服务的主要目标市场是哪里?该服务是否会扩大到其他国家? \\

    Response: 根据文档的内容, Bookertainment 服务的主要目标市场是芬兰。文档提到芬兰目前有约550万居民,并且娱乐行业的就业人数也在增长,这表明Bookertainment服务在芬兰市场存在机会。但是文档也提到, Bookertainment 的最初目标市场是芬兰,\textcolor[rgb]{0.7,0,0}{如果业务 successful},可以逐步扩展到斯堪的纳维亚其他国家。所以服务目前只针对芬兰市场,但未来有可能扩大到斯堪的纳维亚地区。文档没有提及会扩展到其它非斯堪的纳维亚国家。
    \end{CJK} 
    
    \end{tcolorbox}
    \caption{Case study for characteristics of selected samples.}
    \label{fig:rank_case}
\end{figure*}